\documentclass[journal=jacsat,manuscript=article]{achemso}

\usepackage[version=3]{mhchem} 
\usepackage{siunitx}
\usepackage{booktabs}
\usepackage{subcaption}
\usepackage{amssymb}
\usepackage{microtype}
\usepackage{xr}
\usepackage{subcaption}

\author{Zeno Romero}
\affiliation{Laboratory of Engineering Thermodynamics, RPTU Kaiserslautern, Erwin-Schrödinger-Str. 44, 67663 Kaiserslautern, Germany}
\author{Maximilian Kohns}
\affiliation{Laboratory of Engineering Thermodynamics, RPTU Kaiserslautern, Erwin-Schrödinger-Str. 44, 67663 Kaiserslautern, Germany}
\author{Fabian Jirasek}
\affiliation{Laboratory of Engineering Thermodynamics, RPTU Kaiserslautern, Erwin-Schrödinger-Str. 44, 67663 Kaiserslautern, Germany}
\email{fabian.jirasek@rptu.de}

\title[]
  {Predicting Activities in Aqueous Electrolyte Solutions with Hybrid Machine Learning}

\abbreviations{}

\begin{document}







\begin{abstract}

Activities in aqueous electrolyte solutions, usually described by ionic activity and osmotic coefficients, are important properties for modeling many processes in industry and nature. Established activity models, such as those of Pitzer or Bromley, require fitting to experimental data for each electrolyte of interest and thus cannot predict properties for unstudied systems. While some predictive approaches exist, they are typically limited in scope and rely on additional ion-specific descriptors. In this work, we introduce a new hybrid model that combines the physics-based Bromley model with a matrix completion method (MCM) from machine learning. The MCM is employed to predict the electrolyte-specific parameters of the Bromley model, exploiting the fact that these parameters can be arranged in a matrix with cations and anions as rows and columns, respectively. Due to the lack of experimental data for many electrolytes, the initial parameter matrix is sparsely populated, making the prediction of the Bromley parameters for unstudied electrolytes a matrix completion problem. The hybrid model, Bromley-MCM, was trained end-to-end on experimental data for mean ionic activity coefficients and osmotic coefficients of aqueous solutions of 478 electrolytes at 298 K from the Dortmund Data Bank. As output, we obtain a completed matrix of Bromley parameters for 83 cations and 111 anions, enabling consistent prediction of concentration-dependent activities in aqueous solutions of 9,213 electrolytes at 298~K. This substantially extends the applicability of the Bromley model while maintaining high predictive accuracy, as demonstrated through evaluations on electrolytes excluded from model training.

\end{abstract}

\section{Introduction}

Activities in electrolyte solutions play an important role in addressing many critical engineering challenges, e.g.,~electrochemical energy storage \citep{Hei2024} and desalination applications \citep{Keller2021}. Here, the term "activities" is used collectively for different properties that essentially describe the chemical potentials of the ions, the electrolyte, and the solvent in electrolyte solutions. Most established experimental methods for determining such properties focus on solvent activity, often expressed as the osmotic coefficient. These approaches include the isopiestic method\cite{Bousfield1918} and the measurement of the vapor pressure \cite{Robinson1934}, the freezing-point depression\cite{Young1933}, and the osmotic pressure\cite{Heintz2017}. Having conducted a measurement series with either of these methods, the activity of the electrolyte in solution, usually represented in terms of the mean ionic activity coefficient, can be inferred by evaluating the Gibbs-Duhem equation. A notable direct method for determining the mean ionic activity coefficient is the measurement of open-circuit voltages (formerly also called electromotive force measurements)\cite{Argersinger1954}. Many of the aforementioned methods suffer from slow equilibration and reliance on precise references, standards, and/or calibrations\cite{Thomsen2009}. Moreover, these experiments are generally time-consuming and cost-intensive. Thus, it is unfeasible to determine the activity coefficients in all relevant systems experimentally, making predictive methods essential for engineering practice.

Several physical models for activities in electrolyte solutions, such as the excess Gibbs energy ($G^{\rm{E}}$) models of Pitzer\cite{Pitzer1973} and Bromley\cite{Bromley1973}, have been proposed in the literature. These models can reliably generalize across concentration, but their adjustable parameters must be determined by fitting to experimental data for the specific electrolyte, solvent, and temperature of interest. As a consequence, these models cannot generalize over electrolytes, i.e., they cannot be applied to electrolytes for which no activity data are available. This limitation has partially been addressed with the development of LIQUAC and LIFAC\cite{liquacoriginal, modliquac, liquac*}, which enable extrapolations to unstudied temperatures and (mixed) solvents, but still require electrolyte-specific cation-anion interaction parameters that have been fit to data for each electrolyte of interest.  

Only a few models that enable predictions for solutions of completely unstudied electrolytes have been proposed in the literature. Two examples are the ion-specific Bromley model (IBM)\cite{Bromley1973}, which targets the prediction of electrolyte-specific parameters of the Bromley model based on ion-specific parameters, and the Simoes model \cite{Simoes2016}, which, similarly, aims at predicting the electrolyte-specific parameters of the Pitzer model based on ion-specific parameters. However, these approaches have limited predictive accuracy and, in the case of the Simoes model \cite{Simoes2016}, limited applicability, as the ionic radii required as descriptors of the model are not comprehensively available for every ion.

In recent years, machine learning (ML) has emerged as a powerful alternative to and complement of established, physics-based thermodynamic modeling approaches \cite{Jirasek2021,Hoffmann2025}. While purely data-driven ML approaches often lack physical interpretability and robustness, hybrid models that combine established thermodynamic frameworks with data-driven ML models enable generalization across chemical spaces inaccessible to traditional models, effectively closing gaps in established physical models, while retaining thermodynamic consistency \cite{Jirasek2023b,Hoffmann2025b,Specht2024,Hoffmann2026,Hasse2026}. One interesting class of ML methods in thermodnyamics are matrix completion methods (MCMs) \cite{Jirasek2020}, and, by extension, tensor completion methods \cite{Romero2026}, which have also been combined with physical models and knowledge to yield powerful hybrid approaches for predicting various thermodynamic properties, including activity coefficients \cite{Jirasek2020b, Damay2021, Gond2025, Hayer2025cosmo, Zenn2025,Damay2023}, Henry's-law constants \cite{Hayer2022, Hayer2024}, diffusion coefficients \cite{Grossmann2022,Romero2025,Romero2026}, and densities of aqueous electrolyte solutions \cite{KOHNS2025}. Furthermore, MCMs have been embedded in physical modeling frameworks, enabling prediction of model parameters and thereby significantly enhancing their applicability \cite{Jirasek2022, Hayer2025, Hoffmann2024, Jirasek2023b, Jirasek2023, Hayer2025}.

In this work, we present a novel hybrid model for predicting activities in aqueous electrolyte solutions by combining the physics-based Bromley model\cite{Bromley1973} with an MCM. Within this new hybrid model, which we call Bromley-MCM, the MCM is trained to predict the electrolyte-specific adjustable parameters of the Bromley equation, $B_{CA}$, by exploiting the fact that these Bromley parameters for a set of electrolytes can be conveniently arranged in the shape of a matrix, where the rows represent the different cations $C$, the columns represent the different anions $A$, and the entries of the matrix contain the numeric values of $B_{CA}$. Since experimental data to which $B_{CA}$ can be directly fitted exist only for some electrolytes, i.e., only a small fraction of all possible cation-anion pairs $CA$, the prediction of the missing Bromley parameters becomes a matrix completion problem. 

To benchmark the developed Bromley-MCM, we use a consolidated database of mean ionic activity coefficients and osmotic coefficients in aqueous electrolyte solutions at 298~$\pm$~1~K, obtained by careful evaluation of data from the Dortmund Data Bank (DDB)\cite{ddb}. We thereby systematically compare its results with the predictive approaches in the literature, namely, the IBM\cite{Bromley1973} and the model of Simoes \cite{Simoes2016}, each within its respective scope. 

\section{Material and Methods}

\subsection{Electrolyte Activities}

In this work, we address the prediction of activities in aqueous solutions containing a single electrolyte. We thereby assume complete dissociation of the electrolytes in solution following, for an electrolyte $C_{\nu_C} A_{\nu_A}$ that consists of cation $C$ and anion $A$, the reaction
\begin{equation}
C_{\nu_C} A_{\nu_A} \rightarrow \nu_C C^{z_C} + \nu_A A^{z_A},
\label{eq:diss}
\end{equation}
where $\nu_i$ is the stoichiometric coefficient and $z_i$ is the charge number of ion $i$. For brevity, we use $CA$ to denote the electrolyte in the following, omitting the stoichiometric coefficients. To denote solution composition, we use the overall molality $\tilde{m}$ of the electrolyte, which is the number of moles of electrolyte per kilogram of pure water W as the solvent. For complete dissociation, the molality of an ion $m_i$ in the solutions is related to the overall molality $\tilde{m}$ via the mass balance
\begin{equation}
m_i=\nu_i\tilde{m}.
\label{eq:massbalance}
\end{equation}
We furthermore use the ionic strength $I$ of the solution defined as
\begin{equation}
    I = \frac{z_C^2 m_C + z_A^2 m_A}{2 m_0},
\end{equation}
where the division by $m_0=1\mathrm{\:mol\: kg^{-1}}$ is introduced to render the ionic strength dimensionless.
As is common for electrolyte solutions, we consider the chemical potentials of the ions in solution normalized according to Henry's law on the molality scale
\begin{equation}
\mu_{i}(T,p,\tilde{m})=\mu_i^{\mathrm{ref}}(T,p,\tilde{m}\rightarrow 0)+RT\ln{\frac{m_i}{m_0}}+RT\ln{\gamma_i^*(T,p,\tilde{m})},
\label{eq:chempotion}
\end{equation}
where $\mu_i^{\mathrm{ref}}(T,p,\tilde{m}\rightarrow 0)$ is the chemical potential of ion $i$ in the reference state (infinite dilution of the ion), $T$ is the thermodynamic temperature, $p$ is the pressure, $R$ is the universal gas constant, and $\gamma_i^*$ is the activity coefficient of $i$ on the molality scale. Combining these equations for both respective ions yields the overall chemical potential of the electrolyte $CA$, which in case of complete dissociation can be written as
\begin{equation}
\tilde{\mu}_{CA}(T,p,\tilde{m})=\tilde{\mu}_{CA}^{\mathrm{ref}}(T,p,\tilde{m}\rightarrow 0)+ RT\ln{\nu_C^{\nu_C} \nu_A^{\nu_A} \left( \frac{\tilde{m}}{m_0} \right)^\nu}+\nu RT\ln{\tilde{\gamma}_\pm^*(T,p,\tilde{m})},
\label{eq:chempotsalt}
\end{equation}
where
\begin{equation}
    \tilde{\gamma}_{\pm}^* = \left( \left({\gamma_C^*}\right)^{\nu_C}\ \left({\gamma_A^*}\right)^{\nu_A}\right)^{\frac{1}{\nu}} 
    \label{eq:miac}
\end{equation}
is the so-called mean ionic activity coefficient of the electrolyte and 
\begin{equation}
    \nu = \nu_C + \nu_A.
    \label{eq:nu}
\end{equation}
For the chemical potential of the solvent water W, we use the normalization according to Raoult
\begin{equation}
    \tilde{\mu}_W(T,p,\tilde{m})=\tilde{\mu}_W^{\mathrm{pure\ liq}}(T,p)-\nu M_W \tilde{m} RT \phi(T,p,\tilde{m}),
    \label{eq:chempotsolvent}
\end{equation}
where $\tilde{\mu}_\mathrm{W}^{\mathrm{pure\ liq}}(T,p)$ is the chemical potential of pure water W, $M_\mathrm{W}$ is the molar mass of water, and $\phi$ is the osmotic coefficient. Again, equation (\ref{eq:chempotsolvent}) holds for the case of complete dissociation of the electrolyte in the solvent.

In the following, we use $\tilde{\gamma}_{\pm}^*$ and $\phi$ as measures for the activity of the electrolyte and the solvent water in aqueous electrolyte solutions, respectively.

\subsection{Database}

Raw activity data for $\tilde{\gamma}_{\pm}$ and $\phi$ were taken from the DDB 2026 "ELE" database (Vapor-Liquid Equilibria of Electrolyte Systems)\cite{ddb} for binary systems containing exactly one electrolyte and the solvent water. We restricted the scope to these systems because the vast majority of available activity data for aqueous solutions is at 298~K, with very little data at other temperatures. Specifically, we found 92,254 data points for solutions of single electrolytes in water, whereas only 4,180 data points were available for the solvent methanol, and even fewer for other solvents. For the aqueous systems, 43,920 data points are reported at $298\pm1$~K, whereas substantially fewer data points are available at any other temperature. Further details on the data availability are reported in Figures S1 and S2 in the Supporting Information. Since the effect of pressure on liquid-phase properties is generally small, especially at low to moderate pressures, and data at elevated pressures are extremely scarce, the influence of pressure was neglected in this study. All data in this study were either taken at ambient pressure or at the solvent's vapor pressure. 

The following data consolidation and filtering procedure was applied to the raw data from the DDB; further details are reported in the Supporting Information:
\begin{itemize}
    \item Only aqueous systems, i.e., systems with water as the only solvent, were considered.
    \item Electrolytes that could not be split into 1 type of cation and 1 type of anion were excluded. One example of an excluded electrolyte is magnesium ammonium phosphate (MgNH\textsubscript{4}PO\textsubscript{4}), which has two types of cations. On the other hand, ionic coordination complexes, which do not further dissociate in solution, such as tris(ethylenediamine)cobalt(III) ([Co(en)\textsubscript{3}]\textsuperscript{3+}), are included and are treated as single ions. 
    \item Only data reported at $298\pm1$~K were considered.
    \item Only data for which the electrolyte concentration is reported in terms of molality $\tilde{m}$ were used. We also checked the availability of data reported in other concentration measures, but did not find additional relevant systems.
    \item Data points at $I>6$ were excluded (which is reported as the upper bound for the applicability of the Bromley model in the literature\cite{Bromley1973}).
    \item The database was divided into one database for $\tilde{\gamma}_{\pm}$ and one for $\phi$, and each was checked for internal consistency. For this purpose, for a given electrolyte (e.g., NaCl), all available data sets (i.e., data reported by individual sources) in the database were compared. If two data points were reported at the same (to the second decimal) molality, the mean of these values was calculated. However, whenever one of these data points deviated by more than 5 \% from this mean, the entire data set containing this data point was excluded. This procedure was done for $\phi$ and $\tilde{\gamma}_{\pm}$ individually. 
    \item For each electrolyte, the Bromley parameter $B_{CA}$ was fitted individually to the considered $\phi$ and $\tilde{\gamma}_{\pm}$ data for this electrolyte using a least-squares procedure. If the relative mean residual of the fit exceeded 5 \% for either $\phi$ or $\tilde{\gamma}_{\pm}$, the respective electrolyte was excluded from the dataset. This procedure was used since weak electrolytes, for which the Bromley model is not suited, were not excluded a priori based on their respective solubility products. Instead, only those electrolytes for which the Bromley model reproduced the available experimental data with sufficient accuracy (as specified above) were retained. 
    \item Finally, for the retained electrolytes for which both $\phi$ and $\tilde{\gamma}_{\pm}$ were available, the Bromley model was fit to $\phi-1$ and $\ln\tilde{\gamma}_{\pm}$ simultaneously, and electrolytes with residuals greater than $5 \%$ were excluded. This procedure was used to numerically verify Gibbs-Duhem consistency of the data for these electrolytes (within the specified tolerance), since the Bromley model itself is Gibbs-Duhem consistent.
\end{itemize}

\begin{figure}[H]
    \begin{subfigure}{.49\textwidth}
    \centering
    \caption*{Full database}
    \includegraphics[width=1\textwidth]{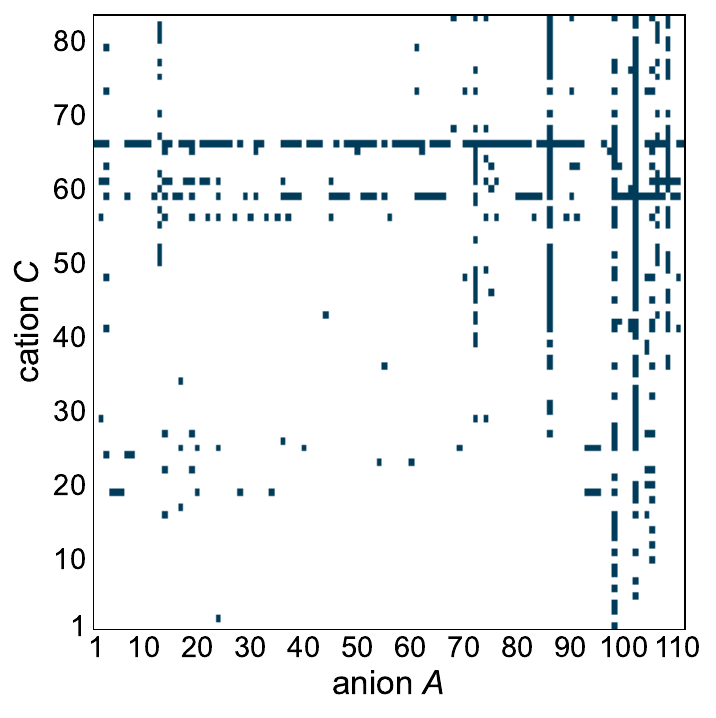}
    \end{subfigure}
    \begin{subfigure}{.49\textwidth}
    \centering
    \caption*{Reduced database}
    \includegraphics[width=1\textwidth]{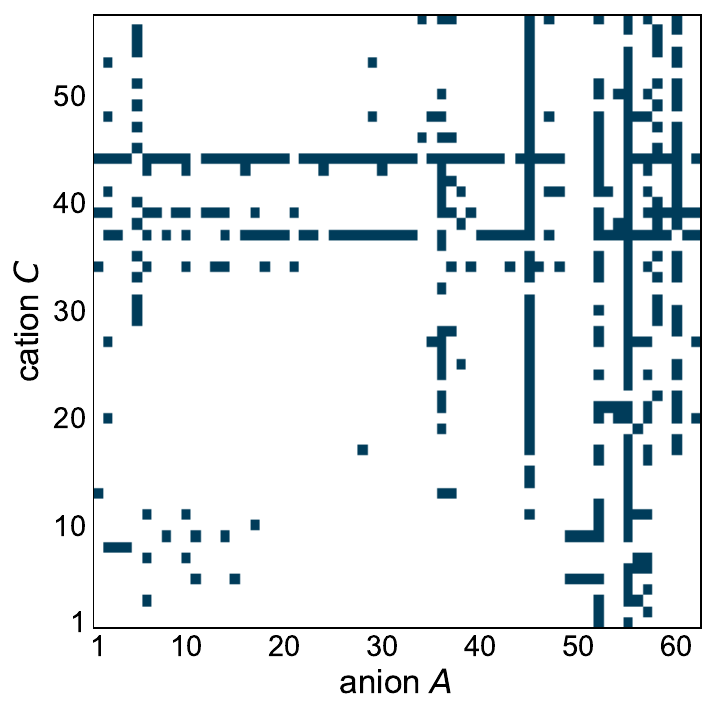}
    \end{subfigure}
    \caption{Experimental data availability (blue cells) for $\tilde{\gamma}_{\pm}$ and/or $\phi$ for cation-anion pairs at $T=298\pm 1$~K in aqueous systems in the full (left) and reduced (right) databases considered in the present work. For the numbers identifying the cations and anions, see Tables S1 and S2 in the Supporting Information, respectively. White cells indicate electrolytes for which no experimental data are available in the databases.} 
    \label{fig:database}
\end{figure}

We call the database containing all data retained after applying the above-described criteria the \textit{full database} in the following, which can be represented as a matrix, where rows correspond to the different cations $C$ and columns to the different anions $A$. This matrix is visualized in Figure~\ref{fig:database} (left). The numbers of cations, anions, available electrolytes, and data points in our full database are listed in Table \ref{tab:stats}, column "Full". The names of all cations and anions are listed in Table S1 in the Supporting Information. The full database contains data for 83 unique cations and 111 unique anions, with experimental mean ionic activity coefficients $\tilde{\gamma}_{\pm}$ and/or osmotic coefficients $\phi$ existing for 478 of those cation-anion pairs ($5.1~\%$ of all possible pairs) with a total of 11,533 experimental data points for $\tilde{\gamma}_{\pm}$ and 13,296 data points for $\phi$. 

To evaluate the predictive accuracy of the hybrid model developed in this work, we use a leave-one-electrolyte-out strategy, in which the model is repeatedly trained on data from all electrolytes except one, whose data are used as test data and compared against the model predictions. This procedure requires data on at least two cation–anion pairs (i.e., different electrolytes) per ion, so we further restricted the full database to include only cations and anions with at least two experimentally available pairs for this analysis. The resulting \textit{reduced database} is shown in Figure \ref{fig:database} (right) in matrix form. The number of cations, anions, and the number of available electrolytes and data points are listed in Table \ref{tab:stats}, column "Reduced". The names of the cations and anions included in the reduced dataset are listed in Table S2 in the Supporting Information. 

\begin{figure}[H]
\centering
    \centering
    \begin{subfigure}{.49\textwidth}
    \includegraphics[width=1\textwidth]{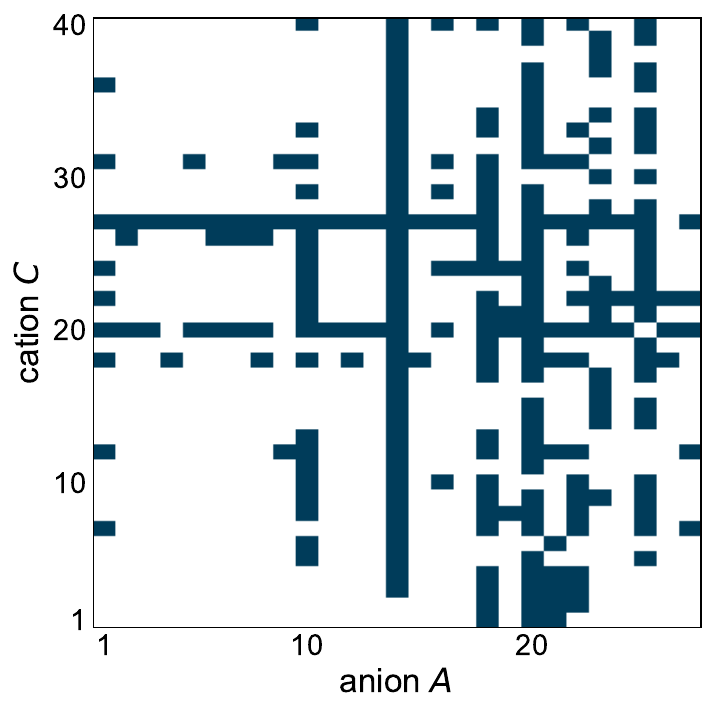}
    \end{subfigure}
    \caption{Experimental data availability for $\tilde{\gamma}_{\pm}$ and/or $\phi$ for cation-anion pairs at $T=298\pm 1$~K in aqueous systems in the Simoes database. For the numbers identifying the cations and anions, see Table S3 in the Supporting Information. White cells indicate electrolytes for which no experimental data are available.} 
    \label{fig:horizons}
\end{figure}

To enable a fair comparison between our new model and the literature model from Simoes \cite{Simoes2016}, we had to define an additional restricted database containing only cation-anion pairs within the scope of the Simoes model. For this database, referred to as the Simoes database below, the same above-described consolidation and filtering procedure was applied, except that no data points were excluded because the Bromley model did not fit them well. The reason for not applying this criterion here is that we did not want to introduce a bias in favor of our hybrid model, which is based on the Bromley model architecture, by considering only data that the original Bromley model can well describe. 
Figure \ref{fig:horizons} visualizes the Simoes database as a matrix. The number of cations, anions, available electrolytes, and data points in these databases are listed in Table \ref{tab:stats}, column "Simoes", respectively.

\begin{table}[H]
\begin{tabular}{llllll}
\toprule
                 & Full          & Reduced     & Simoes           \\
\midrule
Cations          & 83           & 57           & 40                  \\
Anions           & 111          & 62           & 27                    \\
Electrolytes (\%) & 478 (5.1\%) & 401 (11.3 \%) & 256 (23.7 \%)   \\
Data points      & 24,856        & 23,317        & 21,663            \\
\bottomrule
\end{tabular}
\caption{Information on the availability of experimental $\tilde{\gamma}_{\pm}$ and/or $\phi$ data in our full, reduced, and Simoes databases, cf. Figures \ref{fig:database} and \ref{fig:horizons}.} 
\label{tab:stats}
\end{table}

When comparing Figures \ref{fig:database} and \ref{fig:horizons}, as well as the numbers reported in Table \ref{tab:stats}, it becomes evident that the predictive scope of the Simoes model\cite{Simoes2016}, i.e., the number of cation-anion pairs that can be predicted with the model, covers only a small fraction of the full and reduced databases considered in this work. 

\subsection{Electrolyte Activity Models}
\subsubsection{Bromley Model}

The Bromley model, which serves as the basis for the new hybrid model developed in this work, is a semi-empirical $G^\mathrm{E}$ model for electrolyte solutions. It is built on a modified Debye-Hückel limiting law, which is valid at high dilutions of the electrolyte, and adds an empirical extension to describe the deviations from that limiting law observed at higher ionic strengths $I$. It uses a single electrolyte-specific parameter $B_{CA}$, which must be fitted to experimental data for each electrolyte of interest, and then allows for generalization of $\tilde{\gamma}_{\pm}$ and $\phi$ over $I$. The Bromley model is given by the following equations (\ref{eq:bromleyact}) to (\ref{eq:psi})\cite{Bromley1973}
\begin{align}
    \label{eq:bromleyact} 
    \ln \tilde{\gamma}_{\pm} &= - \ln (10)\frac{A_\gamma | z_C z_A | \sqrt{I}}{1+\sqrt{I}} + \ln (10)\frac{(0.06+0.6 B_{CA}) | z_C z_A | I}{(1+cI)^2}+\ln (10)B_{CA}I 
    \\
    \label{eq:bromleyosm}
    \phi &= 1 - \ln (10) \sigma \frac{A_\gamma | z_C z_A | \sqrt{I}}{3} + \ln (10) \psi \frac{(0.06+0.6 B_{CA}) | z_C z_A | I}{2}+\frac{\ln (10)}{2}B_{CA}I   
    \\
    \label{eq:c} 
    c &= \frac{1.5}{|z_C z_A|}
    \\
    \label{eq:sigma} 
    \sigma &= \frac{3}{I^{1.5}}\left[ 1 + \sqrt{I} - \frac{1}{1+\sqrt{I}}-2\ln (1 + \sqrt{I}) \right]
    \\
    \label{eq:psi}
    \psi &= \frac{2}{cI} \left[ \frac{1+2cI}{(1+cI)^2} - \frac{\ln(1+cI)}{cI}     \right] 
\end{align}

In these equations, $A_\gamma$ is the Debye-Hückel constant, which for water at 298~K is $A_\gamma=0.5098$ (since the original Bromley equations use base 10 logarithms), and $z_C$ and $z_A$ are the charges of cation $C$ and anion $A$, respectively. We assume complete dissociation in water for all electrolytes considered in this work, i.e., Bromley's extension for associating ions is not considered.\cite{Bromley1973} 

\subsubsection{Ion-Specific Bromley Model}

We use the ion-specific Bromley model (IBM) \cite{Bromley1973}, which is based on ion-specific instead of electrolyte-specific parameters, as a predictive benchmark in this work. The idea behind IBM is to express the electrolyte-specific parameter $B_{CA}$ as a combination of two cation-specific parameters $B_C$ and $\delta_C$ and two anion-specific parameters $B_A$ and $\delta_A$ according to equation (\ref{eq:ibm})\cite{Bromley1973}
\begin{equation}
    B_{CA} = B_C + B_A + \delta_C \cdot \delta_A
    \label{eq:ibm}
\end{equation}
The ion-specific parameters $B_i$ and $\delta_i$ for aqueous solutions at 298~K were originally determined and published by Bromley in 1973 using a least-squares fit to the previously available $B_{CA}$\cite{Bromley1973}. Since new activity data for aqueous electrolyte systems have been published since then and are now available for a much larger scope of ions, we have determined new ion-specific IBM parameters from our full database using the least-squares method \texttt{minimize} from the \texttt{scipy.optimize} package. Furthermore, we applied the same leave-one-electrolyte-out approach as for our newly developed model to the IBM to compare the predictive performance of both models in a fair way. Further details are given in the computational details section below. Additionally, we compared the fit accuracy (without holding out electrolytes) of the re-parametrized IBM to that of the original IBM \cite{Bromley1973}, as reported in Figure S4 in the Supporting Information, and found that the re-parametrization using new data from the DDB 2026\cite{ddb} leads to higher accuracy on our dataset. 

\subsubsection{Simoes Model}

We use the Simoes model\cite{Simoes2016}, which aims at predicting the parameters of the Pitzer $G^\mathrm{E}$ model\cite{Pitzer1973} for electrolytes for which no suitable fitting data are available, using descriptors of the ions that make up an electrolyte, as an additional predictive benchmark in this work. The underlying Pitzer model\cite{Pitzer1973} is, just like Bromley's model\cite{Bromley1973}, a $G^\mathrm{E}$ model that generalizes over ionic strength $I$, but uses more than one electrolyte-specific adjustable parameter. In its most commonly used form, which is also the basis for the Simoes model\cite{Simoes2016}, three parameters are used ($B_0$, $B_1$, and $C_\phi$). With the Simoes model\cite{Simoes2016}, $B_0$ and $B_1$ can be predicted as functions of the radii $r_i$ and the charge numbers $z_i$ of the ions, while $C_\phi$ is always set to 0. For more details on the Simoes and Pitzer models, we refer to the original publications\cite{Simoes2016,Pitzer1973}. In this work, the Simoes model was implemented as described in the original publication\cite{Simoes2016}, without adjustments, using ionic radii from Marcus\cite{Marcus1994}, which are the same as those used by Simoes et al.\cite{Simoes2016} in the development of their model.

\subsubsection{Hybrid Bromley-MCM}

\begin{figure}[H]
    \centering
    \includegraphics[width=\textwidth]{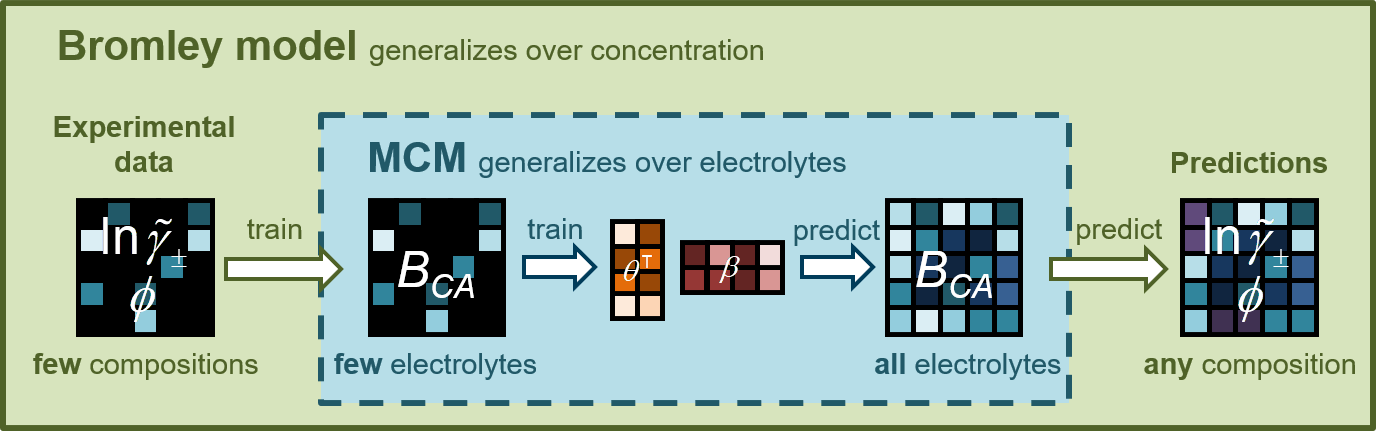}
    \caption{Schematic representation of the hybrid Bromley-MCM for predicting $\tilde{\gamma}_{\pm}$ and $\phi$ in aqueous electrolyte solutions developed in this work. The MCM is embedded in the Bromley model and trained to predict the electrolyte-specific Bromley parameter $B_{CA}$ for unstudied cation-anion pairs, while the Bromley model enables consistent generalization across composition.} 
    \label{fig:bromley-mcm}
\end{figure}

Figure \ref{fig:bromley-mcm} illustrates how the hybrid model introduced in this work, Bromley-MCM, was developed by embedding a matrix completion method (MCM) into the framework of the Bromley model.  

While the \textit{electrolyte-specific} Bromley parameter $B_{CA}$ could directly be fitted only for a small fraction of all electrolytes, given the sparse experimental data for $\tilde{\gamma}_{\pm}$ and $\phi$, the MCM at the core of Bromley-MCM decomposes the matrix of $B_{CA}$ values, spanned by the considered cations and anions, into the product of two low-rank matrices $\pmb{\theta}$ and $\pmb{\beta}$ containing \textit{ion-specific} features. Specifically, each Bromley parameter $B_{CA}$ is thereby modeled as
\begin{equation}
    B_{CA} = \pmb{\theta}_C \cdot \pmb{\beta}_A
    \label{eq:bromley_mcm}
\end{equation}

where $\pmb{\theta}_C$ and $\pmb{\beta}_A$ are the vectors containing the features of cation $C$ and anion $A$, respectively. All feature vectors have length $K$, a hyperparameter of the method set to $K=3$ based on preliminary experiments (cf. Figure S3 in the Supporting Information). 

Bromley-MCM was trained end-to-end on the $\tilde{\gamma}_{\pm}$ and $\phi$ data from our dataset, thereby directly inferring the ion-specific features stored in $\pmb{\theta}$ and $\pmb{\beta}$. In theory, both $\tilde{\gamma}_{\pm}$ and $\phi$ contain the same information concerning $B_{CA}$; we have nevertheless decided to train on both, since for many electrolytes, only one property is available in our database (cf. Figure S8 in the Supporting Information). 

Training was performed in a Bayesian framework, which treats all data and parameters as random variables following probability distributions. Similar Bayesian frameworks are described in detail in our previous work\cite{Jirasek2020, Jirasek2020b, Damay2021, Gond2025, Hayer2025cosmo, Zenn2025,Damay2023,Hayer2022, Hayer2024,Grossmann2022,Romero2025,Romero2026,Jirasek2022, Hayer2025, Hoffmann2024, Jirasek2023b, Jirasek2023, KOHNS2025}; we therefore keep the description in the following brief and focused on the specifics relevant to this work. 

The goal of Bayesian inference is to find the so-called posterior distribution, i.e., the probability distribution of the model parameters ($\pmb{\theta}$ and $\pmb{\beta}$ here) based on the data ($\tilde{\gamma}_{\pm}$ and $\phi$ here). The posterior was calculated by defining a generative model for the data using two further probability distributions: the prior, which describes the probability of the parameters before training on the data, and the likelihood, which describes the conditional probability of the data given the parameters. In this work, Cauchy distributions centered around $\mu_0 = 0$ with scale $\sigma_0=0.2$ were chosen as prior for all features $\pmb{\theta}$ and $\pmb{\beta}$. As the likelihood, each observed value of $\tilde{\gamma}_{\pm}$ or $\phi$ was modeled using a Cauchy distribution with scale parameter $\lambda=0.1$, centered at the corresponding model prediction. These predictions were calculated from the Bromley equations (\ref{eq:bromleyact}) - (\ref{eq:psi}), using $B_{CA}=\pmb{\theta}_C\cdot\pmb{\beta}_A$.

Based on the prior and likelihood, the posterior was calculated using Bayes' law, yielding a probability distribution for each ion-specific feature informed by the training data. Using the mean of these distributions, the electrolyte-specific Bromley parameters $B_{CA}^\mathrm{pred}$ were predicted for all electrolytes that can be built with the studied ions, following equation (\ref{eq:bromley_mcm}). Finally, the $B_{CA}^\mathrm{pred}$ were used for the prediction of $\tilde{\gamma}_{\pm}^\mathrm{pred}$ and $\phi^\mathrm{pred}$ using equations (\ref{eq:bromleyact}) - (\ref{eq:psi}), and compared to the respective experimental data, if available. The prediction error of Bromley-MCM is evaluated using leave-one-electrolyte-out analysis, as explained in the computational details section. 

We report complete sets of $B_{CA}^\mathrm{pred}$ parameters, obtained by training on our full database in the Supporting Information. Although Bromley-MCM was trained on data for electrolytes that can be (approximately) assumed fully dissociated in solution at the studied conditions, predictions for weak electrolytes are also generated, since some combinations of ions making up two strong electrolytes in our database can form weak electrolytes. As one example, consider sodium acetate (NaAc) and hydrochloric acid (HCl), both strong electrolytes in our database (cf. Figure \ref{fig:database}), but acetic acid (HAc) is a weak electrolyte. Predicted $B_{CA}^\mathrm{pred}$ for such electrolytes should thus be used with care, since the Bromley model is not capable of correctly modeling their $\tilde{\gamma}_{\pm}$ and $\phi$ behavior at higher $I$. Such weak electrolytes include many carboxylic acids, phosphoric acid, arsenic acid, transition-metal oxoacids, and non-quaternary amines. We have marked such electrolytes directly in the tabulated $B_{CA}$ in the Supporting Information.

\subsection{Computational Details and Evaluation}

The hybrid Bromley-MCM was implemented in the probabilistic programming language Stan\cite{stan} using its Python package \texttt{CmdStanPy}. Bayesian inference was performed using automatic differentiation variational inference\cite{Kucukelbir2017,Blei2017}. 

The predictive performance of Bromley-MCM and the IBM \cite{Bromley1973} was evaluated using a leave-one-electrolyte-out analysis \cite{Cawley2003}, whereby the models were trained on a subset of the experimental $\tilde{\gamma}_{\pm}$ and $\phi$ data, including all available experimental data points except those of one "test" electrolyte $CA$ to be predicted. By this leave-one-out procedure, predictions for $\tilde{\gamma}_{\pm}$ and $\phi$ were made for truly unseen electrolytes. To evaluate the prediction accuracy of the models, we computed the absolute prediction errors, which, for the predicted quantities $y \in \left\{ \phi, \ln\tilde{\gamma}_{\pm} \right\}$, are defined as:
\begin{equation}
    \mathrm{AE}_i = | y^\mathrm{pred}_i - y^\mathrm{exp}_i |
    \label{eq:absolute_error}
\end{equation}
for each test data point $i$. These pointwise values were aggregated across all experimental data points for a specific electrolyte and are reported as box plots below.

Since our final Bromley-MCM uses $K=3$, i.e., three parameters per ion, and the IBM uses only two parameters per ion, we also report prediction errors for a Bromley-MCM version where we set $K=2$ to enable a clean comparison of the two model architectures that does not depend on the number of trainable parameters. 

For the comparison of Bromley-MCM to the Simoes model\cite{Simoes2016}, the leave-one-electrolyte results were used for Bromley-MCM, guaranteeing true predictions for unseen electrolytes. By contrast, the Simoes model \cite{Simoes2016} was used in their latest published versions, including the reported parameters, so that the results for these benchmark models are, for some electrolytes, not true predictions but correlations.

\section{Results and Discussion}

\subsection{Results of Bromley-MCM and Comparison to IBM}

\begin{figure}[H]
    \centering
    \includegraphics[width=0.6\textwidth]{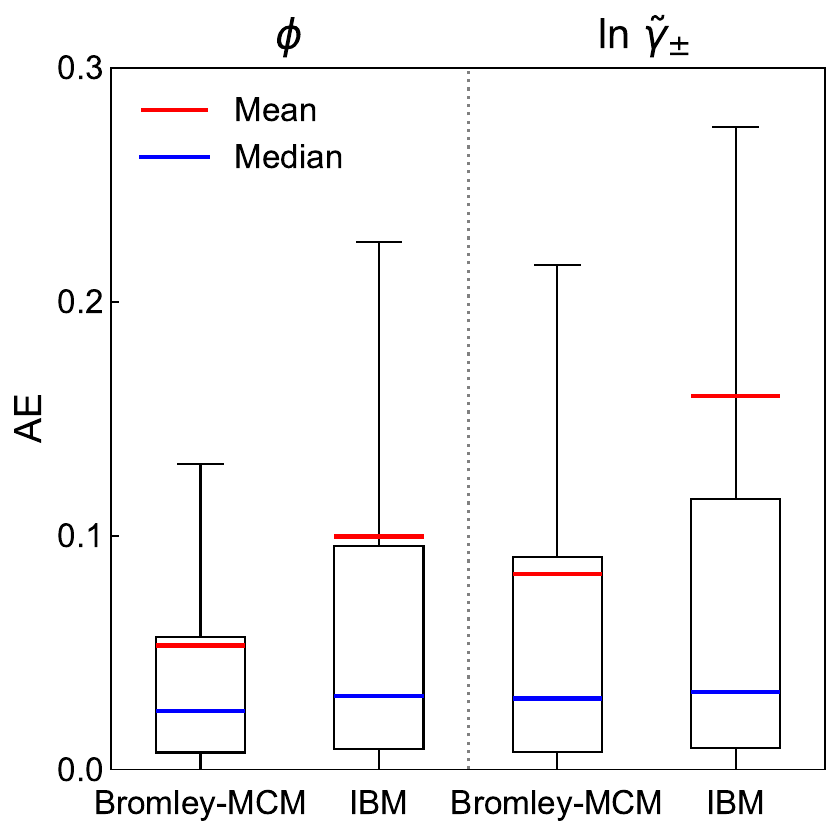}
    \caption{Box plot of the $\ln \tilde{\gamma}_{\pm}$ and $\phi$ prediction errors of Bromley-MCM and IBM on the reduced database evaluated using leave-one-electrolyte-out analysis. Boxes represent interquartile range (IQR), and whiskers denote 1.5 IQR.}
    \label{fig:ibmvsbromley}
\end{figure}

Figure \ref{fig:ibmvsbromley} shows a box plot of the prediction errors of $\ln \tilde{\gamma}_{\pm}$ and $\phi$ with Bromley-MCM and IBM, retrained on the same data, on our reduced database (cf. Figure \ref{fig:database}). All results are for unseen electrolytes. The results show that the new Bromley-MCM substantially outperforms the retrained IBM for both predicted properties. 

For both models, we observe higher prediction errors for $\ln \tilde{\gamma}_{\pm}$ than for $\phi$. Given that the original Bromley model -- the physical model underlying both approaches -- is Gibbs-Duhem-consistent, this is surprising at first glance. This discrepancy may reflect lower quality in the reported $\tilde{\gamma}_{\pm}$ data. In many cases, $\tilde{\gamma}_{\pm}$ are derived from $\phi$ data through Gibbs-Duhem integration. This conversion is particularly sensitive in the low-concentration region, because the infinite-dilution limit provides the integration boundary and small errors in $\phi$ can propagate into $\ln \tilde{\gamma}_{\pm}$. Since highly precise measurements in this regime are difficult, some inconsistencies may have been introduced into the reported $\tilde{\gamma}_{\pm}$ values from the original publications that were not detected by our Gibbs-Duhem verification during data consolidation.

We additionally provide the fit residuals of Bromley-MCM and IBM (Figure S5) and the prediction errors for a Bromley-MCM with $K=2$ (i.e., the same number of parameters as the IBM, Figure S6) in the Supporting Information, all of which show the same qualitative results. 

\subsection{Comparison of Bromley-MCM to the Simoes Model}

\begin{figure}[H]
    \centering
    \includegraphics[width=0.6\textwidth]{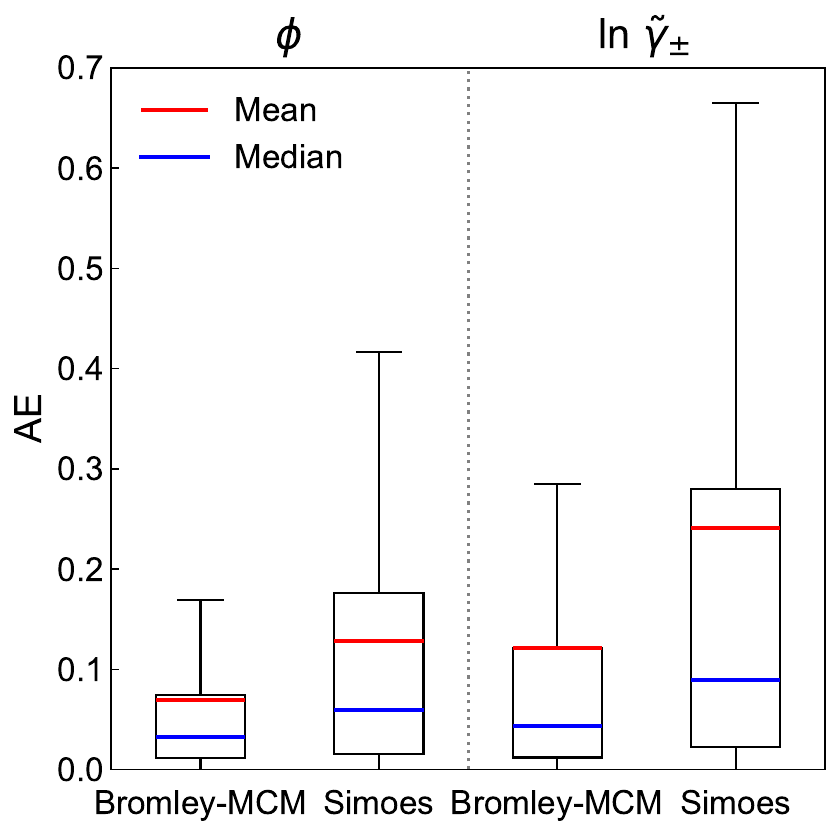}
    \caption{Box plot of the $\ln \tilde{\gamma}_{\pm}$ and $\phi$ prediction errors on the Simoes horizon for Bromley-MCM and the Simoes model. Bromley-MCM was evaluated by leave-one-electrolyte-out analysis, whereas the parameters of the Simoes model were adopted from the original literature\cite{Simoes2016}. Boxes represent IQR and whiskers denote 1.5 IQR.}
    \label{fig:simoesvsbromley}
\end{figure}

Figure \ref{fig:simoesvsbromley} shows the box plot of the prediction errors of $\ln \tilde{\gamma}_{\pm}$ and $\phi$ for Bromley-MCM and Simoes on the Simoes horizon (cf. Figure \ref{fig:horizons}). The results of Bromley-MCM are true predictions for unseen electrolytes. In contrast, the Simoes model was adopted directly from the literature \cite{Simoes2016}, so its results are partly correlations rather than true predictions. In Figure S7 in the Supporting Information, we report the prediction error of the Simoes model on the seen and unseen electrolytes separately and find similar results to those discussed below. 

Figure \ref{fig:simoesvsbromley} shows that Bromley-MCM substantially outperforms the Simoes model\cite{Simoes2016} both on the prediction of the logarithmic mean ionic activity coefficient and the osmotic coefficient. The Simoes model shows large outliers and a very large interquartile range. Our Bromley-MCM not only predicts aqueous electrolyte activities more accurately than the established Simoes model \cite{Simoes2016}, but also covers a substantially broader chemical space, enabling predictions for more than 9,000 electrolytes.

\subsection{Predicted Electrolyte Activities as a Function of Ionic Strength}

\begin{figure}[H]
\centering
    \captionsetup[subfigure]{
        justification=centering,
        singlelinecheck=false,
        margin={4mm,-4mm}
    }
    
    \begin{subfigure}{.49\textwidth}
    \centering
    \caption*{Calcium chloride (CaCl\textsubscript{2})}
    \includegraphics[width=1\textwidth]{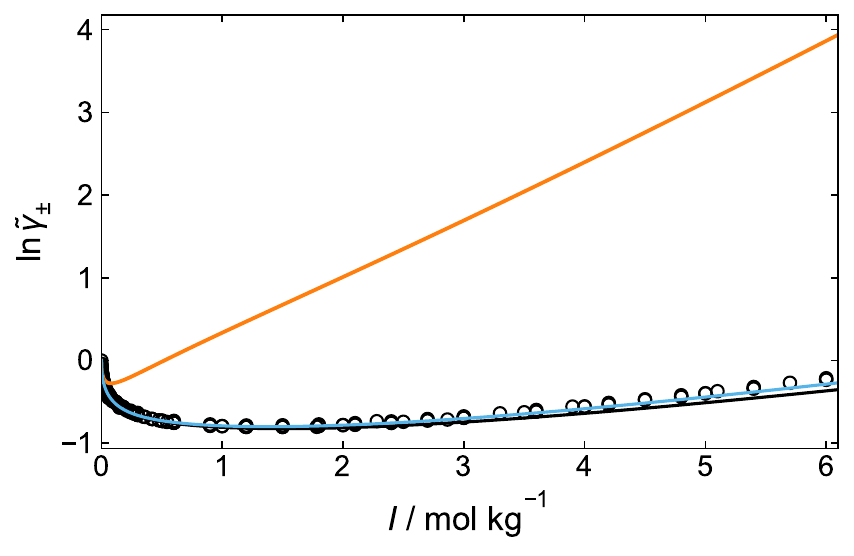}
    \end{subfigure}
    \begin{subfigure}{.49\textwidth}
    \centering
    \caption*{Potassium tosylate (KTos)}
    \includegraphics[width=1\textwidth]{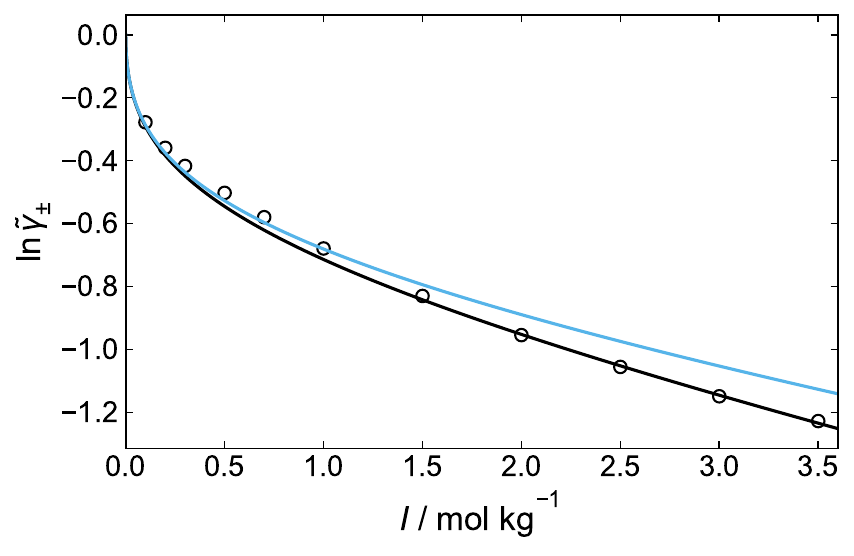}
    \end{subfigure}
    \begin{subfigure}{.49\textwidth}
    \centering
    \caption*{Lutetium ethyl sulfate (Lu(EtSO\textsubscript{4})\textsubscript{3})}
    \includegraphics[width=1\textwidth]{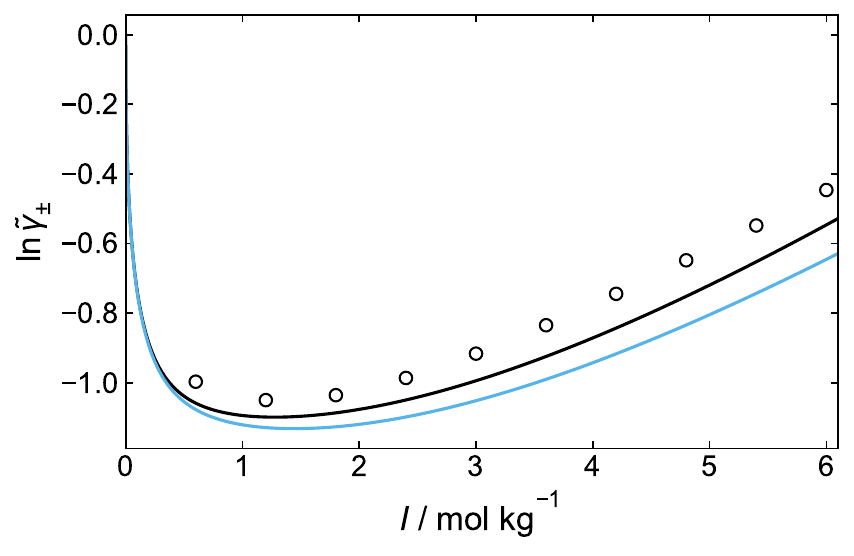}
    \end{subfigure}
    \begin{subfigure}{.49\textwidth}
    \centering
    \caption*{Magnesium bromide (MgBr\textsubscript{2})}
    \includegraphics[width=1\textwidth]{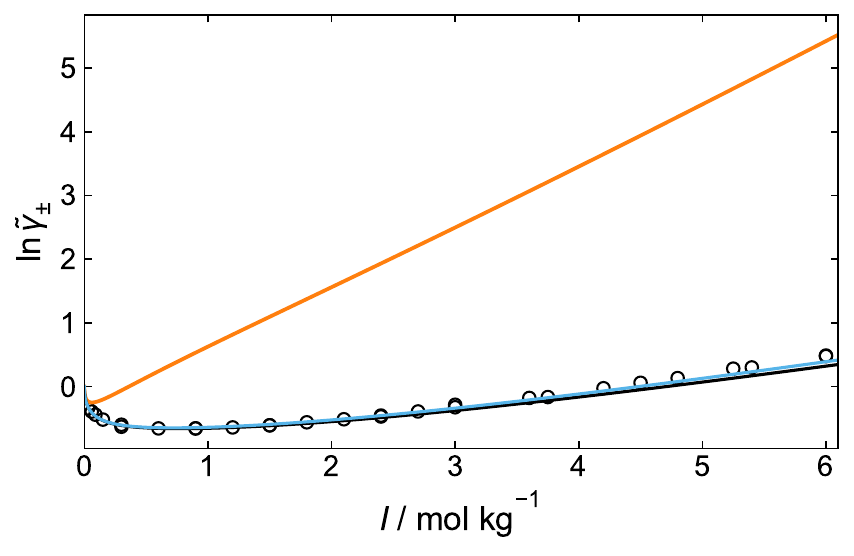}
    \end{subfigure}
    \begin{subfigure}{.49\textwidth}
    \centering
    \caption*{Sodium nitrate (NaNO\textsubscript{3})}
    \includegraphics[width=1\textwidth]{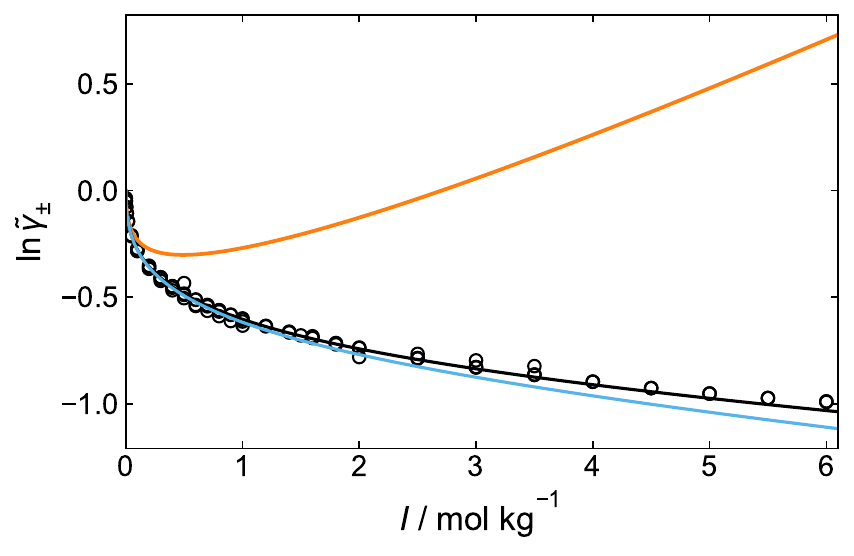}
    \end{subfigure}
    \begin{subfigure}{.49\textwidth}
    \centering
    \caption*{Ammonium chloride (NH\textsubscript{4}Cl)}
    \includegraphics[width=1\textwidth]{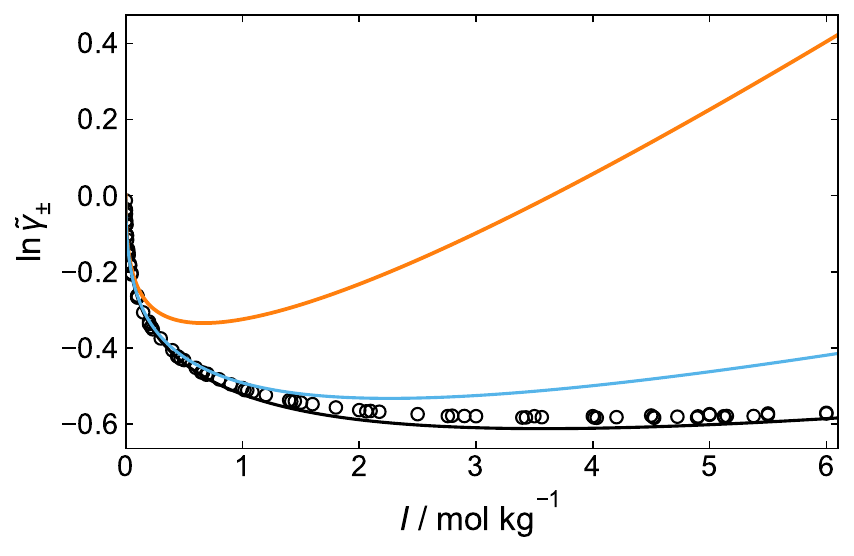}
    \end{subfigure}
    \begin{subfigure}{\textwidth}
    \centering
    \includegraphics[width=0.75\textwidth]{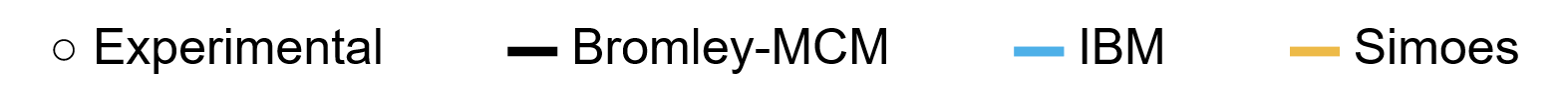}
    \end{subfigure}
    \caption{Prediction of $\ln\tilde\gamma_\pm$ for exemplary electrolytes in aqueous solutions at 298~K as a function of ionic strength. Lines: Predictions with different models. Circles: experimental data from DDB\cite{ddb}.}
    \label{fig:gammas}
\end{figure}

\begin{figure}[H]
\centering
    \captionsetup[subfigure]{
        justification=centering,
        singlelinecheck=false,
        margin={4mm,-4mm}
    }

    \begin{subfigure}{.49\textwidth}
    \centering
    \caption*{Hydrogen bromide (HBr)}
    \includegraphics[width=1\textwidth]{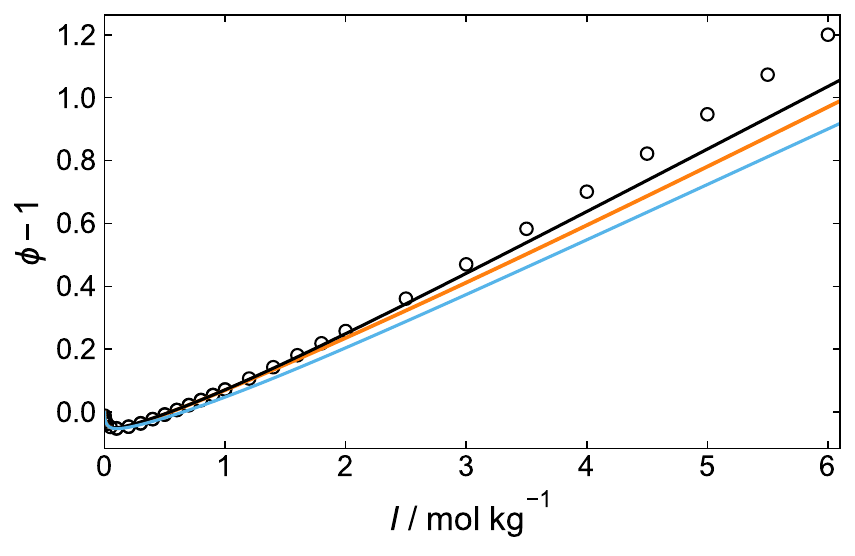}
    \end{subfigure}
    \begin{subfigure}{.49\textwidth}
    \centering
    \caption*{Tris(ethylenediamine)cobalt(III) perchlorate (Co(en)\textsubscript{3}(ClO\textsubscript{4})\textsubscript{3})}
    \includegraphics[width=1\textwidth]{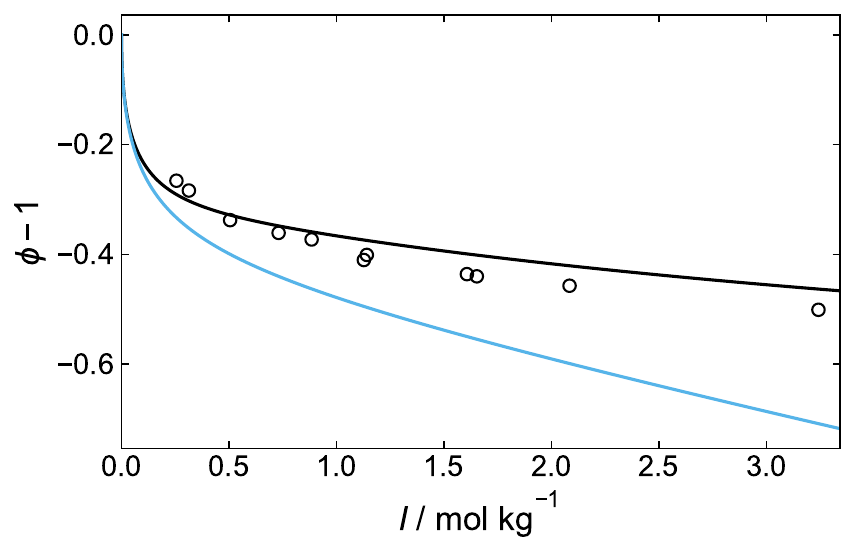}
    \end{subfigure}
    \begin{subfigure}{.49\textwidth}
    \centering
    \caption*{Potassium hydrogen phosphate (K\textsubscript{2}HPO\textsubscript{4})}
    \includegraphics[width=1\textwidth]{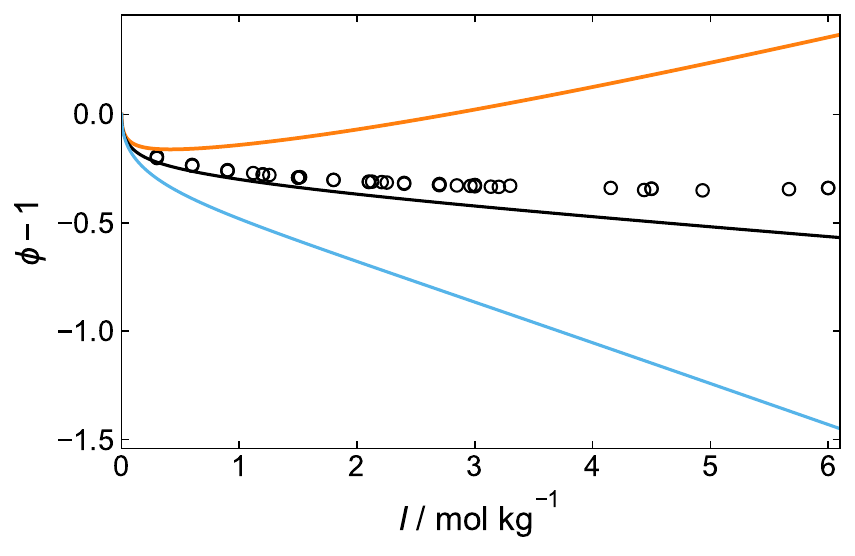}
    \end{subfigure}
    \begin{subfigure}{.49\textwidth}
    \centering
    \caption*{Sodium acetate (NaAc)}
    \includegraphics[width=1\textwidth]{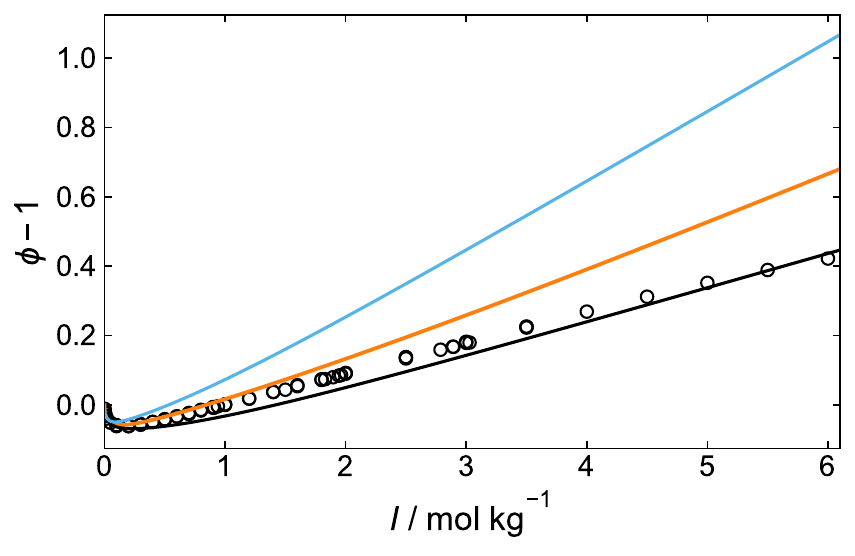}
    \end{subfigure}
    \begin{subfigure}{.49\textwidth}
    \centering
    \caption*{Copper nitrate (Cu(NO\textsubscript{3})\textsubscript{2})}
    \includegraphics[width=1\textwidth]{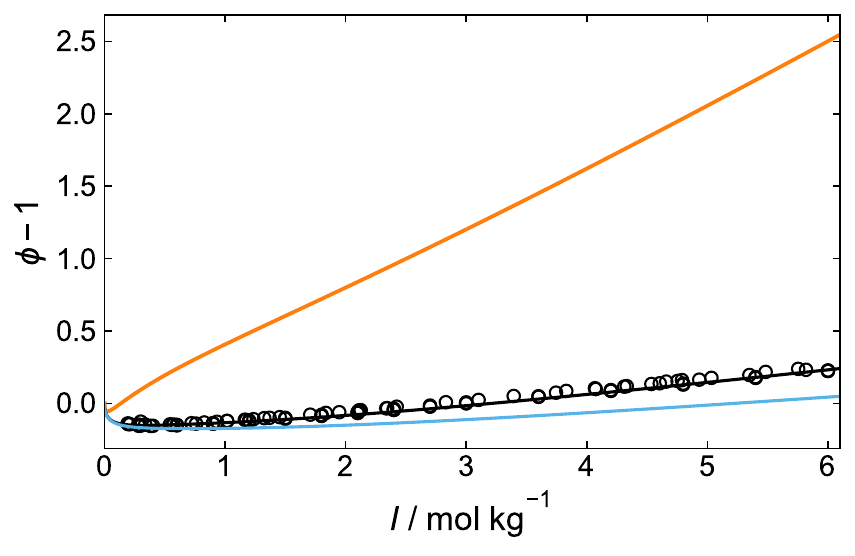}
    \end{subfigure}
    \begin{subfigure}{.49\textwidth}
    \centering
    \caption*{Ammonium chloride (NH\textsubscript{4}Cl)}
    \includegraphics[width=1\textwidth]{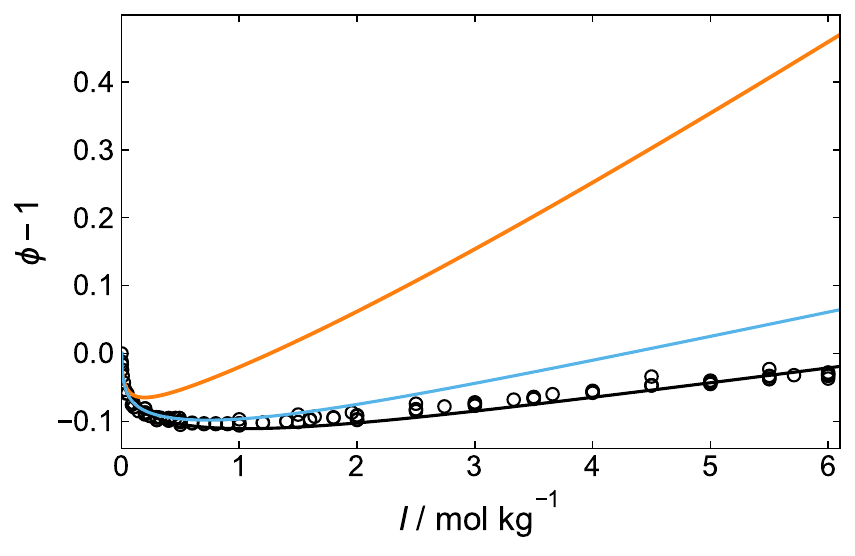}
    \end{subfigure}
    \begin{subfigure}{\textwidth}
    \centering
    \includegraphics[width=0.75\textwidth]{figures/legend.png}
    \end{subfigure}
    \caption{Prediction of $\phi-1$ in exemplary aqueous electrolyte solutions at 298~K as a function of ionic strength. Lines: Predictions with different models. Circles: experimental data from DDB\cite{ddb}.}
    \label{fig:osms}
\end{figure}

Figures \ref{fig:gammas} and \ref{fig:osms} show a comparison between the predictions by the three models studied in this work (Bromley-MCM, IBM, and Simoes) for $\ln \tilde\gamma_\pm$ and $\phi-1$ as a function of ionic strength $I$ and the respective experimental data for several electrolytes as examples. Again, the results of Bromley-MCM and IBM are true predictions for unseen electrolytes, whereas the Simoes model has been fitted to data for most of the shown electrolytes. Wherever lines are missing in a plot panel, the Simoes model could not be applied to the electrolyte under consideration.

The agreement between the experimental data and the predictions with Bromley-MCM is very good for most electrolytes. All curves show the correct qualitative dependence of $\ln \tilde\gamma_\pm$ and $\phi-1$ on $I$, with the characteristic minimum at low ionic strength, which is achieved by building the hybrid model on the physics-based Bromley model. For many electrolytes, good agreement is also found for the other models. However, as already shown in Figures \ref{fig:ibmvsbromley} and \ref{fig:simoesvsbromley}, Bromley-MCM outperforms all other approaches overall.

While the Simoes model\cite{Simoes2016}, despite having been fit to data for all shown electrolytes except Potassium hydrogen phosphate (K\textsubscript{2}HPO\textsubscript{4}) and those where no Simoes curve (orange) is present, fails for certain electrolytes, Bromley-MCM also manages to correctly predict the activities of electrolytes that contain organic anions (e.g., Tos\textsuperscript{-}, Ac\textsuperscript{-}, SCN\textsuperscript{-}), electrolyts containing ions with absolute charges greater than one (e.g., Ca\textsuperscript{2+}, Mg\textsuperscript{2+}, Lu\textsuperscript{3+}, PO\textsubscript{4}\textsuperscript{3-}), or electrolytes that contain ionic complexes (e.g., [Co(en)\textsubscript{3}]\textsuperscript{3+}). 

As can, e.g., be seen in Figure \ref{fig:osms} middle left, the prediction of electrolytes containing ions with multiple protonated forms is somewhat worse than that of other electrolytes. While K\textsubscript{3}PO\textsubscript{4} would be expected to behave like a typical strong electrolyte, the resulting PO\textsubscript{4}\textsuperscript{3-}, like other phosphates, undergoes stepwise protonation and deprotonation. Modeling the activities of such acids with high accuracy is challenging, and alternative models for these systems typically require substantially more adjustable parameters \cite{Clegg1994,Pitzer1976}.

Despite some minor shortcomings, mainly due to the limitations of the underlying Bromley model\cite{Bromley1973}, the results clearly demonstrate that the matrix completion concept is well suited for the prediction of Bromley parameters $B_{CA}$ as the hybrid model resulting from embedding an MCM into the Bromley model substantially outperfoms the existing benchmarks in the prediction of $\tilde\gamma_\pm$ and $\phi$ in aqueous electrolyte-containing systems.

\section{Conclusions}

In this work, we present a novel hybrid approach for predicting activities in aqueous electrolyte solutions at 298 K by embedding a matrix completion method (MCM) from machine learning (ML) within the Bromley model to predict the Bromley parameter $B_{CA}$, enabling the consistent modeling of activity coefficients $\tilde\gamma_\pm$ and osmotic coefficients $\phi$ for unstudied electrolyte systems, thereby addressing the limitation of classical electrolyte models as the ones of Bromley\cite{Bromley1973} and Pitzer \cite{Pitzer1973}. 

The new hybrid model, Bromley-MCM, was trained end-to-end and evaluated on data for $\tilde\gamma_\pm$ and $\phi$ from a consolidated database, based on raw data from the Dortmund Data Bank\cite {ddb}, containing experimental data for aqueous solutions of 478 electrolytes, covering 83 unique cations and 111 unique anions. Bromley-MCM significantly outperforms other models reported in the literature that generalize to unstudied electrolytes, namely, the ion-specific Bromley model (IBM)\cite{Bromley1973} and the Simoes model \cite{Simoes2016}, in prediction accuracy, while also offering a substantially broader scope, covering more ions, than the Simoes model. While the extrapolation to unstudied electrolytes is enabled by the MCM, the physics-based Bromley model underlying the hybrid Bromley-MCM guarantees smooth and consistent generalization over ionic strength. 

With this work, we report a complete set of $B_{CA}$ values for 9,213 electrolytes, which can be used to describe activities in aqueous solution at 298~K. For 8,735 of these electrolytes, no data under these conditions were available in the literature. In future work, the hybrid approach introduced here can be extended to further solvents and the temperature-dependent prediction of activities in electrolyte solutions. 

\begin{acknowledgement}

We gratefully acknowledge financial support by DFG in the frame of the Research Training Group GRK 2908 "Valuable Wastewater (WERA)" (grant number 503479768) and the Priority Program SPP 2363 "Molecular Machine Learning" (grant number 497201843). Furthermore, FJ gratefully acknowledges financial support by DFG in the frame of the Emmy-Noether program (grant number 528649696).

\end{acknowledgement}

\begin{suppinfo}

Supporting Information is available:

\begin{itemize}
\item{Compound identifiers; hyperparameter optimization; further error analysis; and analysis of data availability. (PDF)}
\item{Full Bromley parameter set ($B_{CA}$) for 9,213 electrolytes, from training on the full database. (XLSX)}

\end{itemize}

\end{suppinfo}

\section*{Declaration of generative AI and AI-assisted technologies in the manuscript preparation process}

During the preparation of this work the authors used ChatGPT (OpenAI) in order to improve the language of the manuscript and to assist with code development and debugging. After using this tool/service, the authors reviewed and edited the content as needed and take full responsibility for the content of the published article.

\bibliography{achemso-demo}

\end{document}


\section{Ion Identifiers}
\label{ident}
\setcounter{page}{1}

Tables \ref{tab:ions_full} to \ref{tab:ions_simoes} list the identifiers for cations $C$ and anions $A$, respectively, covered in Figures \ref{fig:database} to \ref{fig:horizons}. 

\begingroup
\setlength{\LTleft}{-20cm plus -1fill}
\setlength{\LTright}{\LTleft}
\begin{spacing}{1.0}

\begin{longtable}{llll}
\caption{Identifiers for cations and anions in the \textit{full} database (cf. Figure \ref{fig:database}).} \\
\hline Cation Nr. & Cation    & Anion Nr. & Anion                    \\ \hline
\endfirsthead

\multicolumn{4}{c}%
{{\tablename\ \thetable{} -- continued from previous page}} \\
\hline Cation Nr. & Cation    & Anion Nr. & Anion                                     \\ \hline 
\endhead

\hline
\endfoot

\hline
\endlastfoot

1  & Tetraisobutylammonium                       & 1   & Metaborate                       \\
2  & Triisobutylmethylphosphonium                & 2   & 4,4'-Bibenzyldisulfonate         \\
3  & Tetra-sec-butylammonium                     & 3   & Acetate                          \\
4  & Trimethyloctylammonium                      & 4   & 2-Amino-3-methylbutanoate        \\
5  & N,N,N-Octylbutyldimethylammonium            & 5   & 2-Amino-4-methylpentanoate       \\
6  & n-Octylammonium                             & 6   & 2-Aminopropanoate                \\
7  & 1-Octylpyridinium                           & 7   & Propanoate                       \\
8  & 1-Octyl-3-methylimidazolium                 & 8   & Butanoate                        \\
9  & 1-Heptyl-3-methylimidazolium                & 9   & Pentanoate                       \\
10 & Triethylhexylammonium                       & 10  & Octanoate                        \\
11 & 1-Hexyl-3-methylimidazolium                 & 11  & Decanoate                        \\
12 & Triethylpentylammonium                      & 12  & Ethyl malonate                   \\
13 & 1-Pentyl-3-methylimidazolium                & 13  & Ethyl sulfate                    \\
14 & Butyltriethylammonium                       & 14  & Ethanesulfonate                  \\
15 & Dibutyldiethylammonium                      & 15  & Ethylbenzene sulfonate           \\
16 & Tetrabutylammonium                          & 16  & Methyl malonate                  \\
17 & Tributylmethylphosphonium                   & 17  & Methyl sulfate                   \\
18 & Tributylsulfonium                           & 18  & p-Anisole sulfonate              \\
19 & 1-Butyl-3-methylimidazolium                 & 19  & Methanesulfonate                 \\
20 & Tetrapropylammonium                         & 20  & Thymine                          \\
21 & 1-Propyl-3-methylimidazolium                & 21  & 2,4,6-Trimethylbenzene sulfonate \\
22 & Tetraethylammonium                          & 22  & 2,5-Dimethylbenzenesulfonate     \\
23 & Triethylammonium                            & 23  & 2,4-Dimethylbenzenesulfonate     \\
24 & Diethylammonium                             & 24  & 4-Toluenesulfonate               \\
25 & 1-Ethyl-3-methylimidazolium                 & 25  & 3-Methylsalicylate               \\
26 & Tetramethylguanidinium                      & 26  & Cefazolin                        \\
27 & Tetramethylammonium                         & 27  & Tetrachloroaurate                \\
28 & 2-Hydroxyethyl-trimethylammonium            & 28  & Tetrafluoroborate                \\
29 & Trimethyl-(2-trimethylammonioethyl)ammonium & 29  & Hexafluorophosphate              \\
30 & Trimethylammonium                           & 30  & Hexafluoridosilicate             \\
31 & Dimethylammonium                            & 31  & Thiocyanate                      \\
32 & Timethylsulfonium                           & 32  & Aspartate                        \\
33 & 1-(2-Carboxyethyl)-3-methylimidazolium      & 33  & Arginine                         \\
34 & 1,3-Dimethylimidazolium                     & 34  & 2-Aminoacetate                   \\
35 & Hexaaminecobalt(III)                        & 35  & Lysine                           \\
36 & Guanidinium                                 & 36  & Sulfamate                        \\
37 & Uranyl                                      & 37  & Glutamate                        \\
38 & Tetraethanol ammonium                       & 38  & Sulfanilate                      \\
39 & Silver(I)                                   & 39  & Adenosine triphosphate           \\
40 & Aluminium(III)                              & 40  & Adenine                          \\
41 & Barium (II)                                 & 41  & Dihydrogen citrate               \\
42 & Calcium(II)                                 & 42  & Fumarate                         \\
43 & Cadmiun(II)                                 & 43  & Maleate                          \\
44 & Cerium(III)                                 & 44  & Oxalate                          \\
45 & Cobalt(II)                                  & 45  & Trichloracetate                  \\
46 & Tris(ethylenediamine)cobalt(III)            & 46  & Trifluoroacetate                 \\
47 & Chromium(III)                               & 47  & Tartrate                         \\
48 & Cesium(I)                                   & 48  & Hydrogen malonate                \\
49 & Copper(II)                                  & 49  & Malonate                         \\
50 & Dysprosium(III)                             & 50  & Hydrogen citrate                 \\
51 & Erbium(III)                                 & 51  & Hydrogen succinate               \\
52 & Europium(III)                               & 52  & Succinate                        \\
53 & Iron(II)                                    & 53  & Hydrogen adipate                 \\
54 & Iron(III)                                   & 54  & 2-Hydroxyacetate                 \\
55 & Gadolinium(III)                             & 55  & Carbonate                        \\
56 & Proton                                      & 56  & Mellitate                        \\
57 & Holmium(III)                                & 57  & 4-Hydroxybenzoate                \\
58 & Indium(III)                                 & 58  & Benzoate                         \\
59 & Potassium(I)                                & 59  & Salicylate                       \\
60 & Lanthanum(III)                              & 60  & Formate                          \\
61 & Lithium(I)                                  & 61  & Nitrite                          \\
62 & Lutetium(III)                               & 62  & Dihydrogen phosphate             \\
63 & Magnesium(II)                               & 63  & Dihydrogen pyrophosphate         \\
64 & Manganese(II)                               & 64  & Hydrogen phosphate               \\
65 & Ammonium                                    & 65  & Pyrophosphate                    \\
66 & Sodium(I)                                   & 66  & Phosphate                        \\
67 & Neodymium(III)                              & 67  & Trifluoromethane sulfonate       \\
68 & Nickel(II)                                  & 68  & Ethane-1,2-disulfonate           \\
69 & Lead(II)                                    & 69  & Hydrogen sulfate                 \\
70 & Praseodymium(III)                           & 70  & Peroxydisulfate                  \\
71 & Tris(1,3-Diaminopropane)platinum(IV)        & 71  & Dithionate                       \\
72 & Tris(ethylenediamine)plantinum(IV)          & 72  & Sulfate                          \\
73 & Rubidium(I)                                 & 73  & 4-Hydroxybenzenesulfonate        \\
74 & Scandium(III)                               & 74  & m-Benzenedisulfonate             \\
75 & Samarium(III)                               & 75  & 1,5-Naphthalenedisulfonate       \\
76 & Strontium(II)                               & 76  & Benzenesulfonate                 \\
77 & Terbium(III)                                & 77  & Thiosulfate                      \\
78 & Thorium(IV)                                 & 78  & Sulfite                          \\
79 & Thallium(I)                                 & 79  & Benzenesulfinate                 \\
80 & Thulium(III)                                & 80  & Dihydrogen arsenate              \\
81 & Yttrium(III)                                & 81  & Hydrogen arsenate                \\
82 & Ytterbium(III)                              & 82  & Arsenate                         \\
83 & Zinc(II)                                    & 83  & Dichromate                       \\
   &                                             & 84  & Chromate                         \\
   &                                             & 85  & Molybdate                        \\
   &                                             & 86  & Nitrate                          \\
   &                                             & 87  & 4-Nitrophenolate                 \\
   &                                             & 88  & 3-Nitrobenzenesulfonate          \\
   &                                             & 89  & Perrhenate                       \\
   &                                             & 90  & Selenate                         \\
   &                                             & 91  & Pertechnetate                    \\
   &                                             & 92  & Tungstate                        \\
   &                                             & 93  & Xanthine                         \\
   &                                             & 94  & Hypoxanthine                     \\
   &                                             & 95  & Uracil                           \\
   &                                             & 96  & Dodecahydrododecaborate          \\
   &                                             & 97  & Decahydrodecaborate              \\
   &                                             & 98  & Bromide                          \\
   &                                             & 99  & Hexacyanocobaltate(III)          \\
   &                                             & 100 & Hexacyanoferrite(II)             \\
   &                                             & 101 & Hexacyanoferrate(III)            \\
   &                                             & 102 & Chloride                         \\
   &                                             & 103 & Hexachloroplatinate(IV)          \\
   &                                             & 104 & Fluoride                         \\
   &                                             & 105 & Iodide                           \\
   &                                             & 106 & Bromate                          \\
   &                                             & 107 & Chlorate                         \\
   &                                             & 108 & Perchlorate                      \\
   &                                             & 109 & Iodate                           \\
   &                                             & 110 & Hydroxide                        \\
   &                                             & 111 & Tetraphenylborate               

\label{tab:ions_full}
\end{longtable}

\end{spacing}
\endgroup

\newpage

\begingroup
\setlength{\LTleft}{-20cm plus -1fill}
\setlength{\LTright}{\LTleft}
\begin{spacing}{1.0}

\begin{longtable}{llll}
\caption{Identifiers for cations and anions in the \textit{reduced} database (cf. Figure \ref{fig:database}).} \\
\hline Cation Nr. & Cation    & Anion Nr. & Anion                    \\ \hline
\endfirsthead

\multicolumn{4}{c}%
{{\tablename\ \thetable{} -- continued from previous page}} \\
\hline Cation Nr. & Cation    & Anion Nr. & Anion                                     \\ \hline 
\endhead

\hline
\endfoot

\hline
\endlastfoot

1  & 1-Hexyl-3-methylimidazolium                 & 1  & 4,4'-Bibenzyldisulfonate         \\
2  & Butyltriethylammonium                       & 2  & Acetate                          \\
3  & Tetrabutylammonium                          & 3  & Propanoate                       \\
4  & Tributylsulfonium                           & 4  & Butanoate                        \\
5  & 1-Butyl-3-methylimidazolium                 & 5  & Ethyl sulfate                    \\
6  & Tetrapropylammonium                         & 6  & Ethanesulfonate                  \\
7  & Tetraethylammonium                          & 7  & Ethylbenzene sulfonate           \\
8  & Diethylammonium                             & 8  & Methyl sulfate                   \\
9  & 1-Ethyl-3-methylimidazolium                 & 9  & $p$-Anisole sulfonate              \\
10 & Tetramethylguanidinium                      & 10 & Methanesulfonate                 \\
11 & Tetramethylammonium                         & 11 & Thymine                          \\
12 & 2-Hydroxyethyl-trimethylammonium            & 12 & 2,4,6-Trimethylbenzene sulfonate \\
13 & Trimethyl-(2-trimethylammonioethyl)ammonium & 13 & 2,5-Dimethylbenzenesulfonate     \\
14 & Trimethylammonium                           & 14 & 4-Toluenesulfonate               \\
15 & Dimethylammonium                            & 15 & Tetrafluoroborate                \\
16 & Timethylsulfonium                           & 16 & Thiocyanate                      \\
17 & Guanidinium                                 & 17 & Sulfamate                        \\
18 & Uranyl                                      & 18 & Glutamate                        \\
19 & Silver(I)                                   & 19 & Sulfanilate                      \\
20 & Barium (II)                                 & 20 & Adenosine triphosphate           \\
21 & Calcium(II)                                 & 21 & Trichloracetate                  \\
22 & Cadmiun(II)                                 & 22 & Trifluoroacetate                 \\
23 & Cerium(III)                                 & 23 & Hydrogen malonate                \\
24 & Cobalt(II)                                  & 24 & Hydrogen citrate                 \\
25 & Tris(ethylenediamine)cobalt(III)            & 25 & Hydrogen succinate               \\
26 & Chromium(III)                               & 26 & Succinate                        \\
27 & Cesium(I)                                   & 27 & Hydrogen adipate                 \\
28 & Copper(II)                                  & 28 & Carbonate                        \\
29 & Dysprosium(III)                             & 29 & Nitrite                          \\
30 & Erbium(III)                                 & 30 & Dihydrogen phosphate             \\
31 & Europium(III)                               & 31 & Hydrogen phosphate               \\
32 & Iron(II)                                    & 32 & Pyrophosphate                    \\
33 & Gadolinium(III)                             & 33 & Phosphate                        \\
34 & Proton                                      & 34 & Ethane-1,2-disulfonate           \\
35 & Holmium(III)                                & 35 & Peroxydisulfate                  \\
36 & Indium(III)                                 & 36 & Sulfate                          \\
37 & Potassium(I)                                & 37 & $m$-Benzenedisulfonate             \\
38 & Lanthanum(III)                              & 38 & 1,5-Naphthalenedisulfonate       \\
39 & Lithium(I)                                  & 39 & Benzenesulfonate                 \\
40 & Lutetium(III)                               & 40 & Dihydrogen arsenate              \\
41 & Magnesium(II)                               & 41 & Hydrogen arsenate                \\
42 & Manganese(II)                               & 42 & Arsenate                         \\
43 & Ammonium                                    & 43 & Dichromate                       \\
44 & Sodium(I)                                   & 44 & Chromate                         \\
45 & Neodymium(III)                              & 45 & Nitrate                          \\
46 & Nickel(II)                                  & 46 & Perrhenate                       \\
47 & Praseodymium(III)                           & 47 & Selenate                         \\
48 & Rubidium(I)                                 & 48 & Pertechnetate                    \\
49 & Samarium(III)                               & 49 & Xanthine                         \\
50 & Strontium(II)                               & 50 & Hypoxanthine                     \\
51 & Terbium(III)                                & 51 & Uracil                           \\
52 & Thorium(IV)                                 & 52 & Bromide                          \\
53 & Thallium(I)                                 & 53 & Hexacyanocobaltate(III)          \\
54 & Thulium(III)                                & 54 & Hexacyanoferrate(III)            \\
55 & Yttrium(III)                                & 55 & Chloride                         \\
56 & Ytterbium(III)                              & 56 & Fluoride                         \\
57 & Zinc(II)                                    & 57 & Iodide                           \\
   &                                             & 58 & Bromate                          \\
   &                                             & 59 & Chlorate                         \\
   &                                             & 60 & Perchlorate                      \\
   &                                             & 61 & Iodate                           \\
   &                                             & 62 & Hydroxide                                     

\label{tab:ions_reduced}
\end{longtable}

\end{spacing}
\endgroup

\newpage

\begingroup
\setlength{\LTleft}{-20cm plus -1fill}
\setlength{\LTright}{\LTleft}
\begin{spacing}{1.0}

\begin{longtable}{llll}
\caption{Identifiers for cations and anions in the Simoes database (cf. Figure \ref{fig:horizons}).} \\
\hline Cation Nr. & Cation    & Anion Nr. & Anion                    \\ \hline
\endfirsthead

\multicolumn{4}{c}%
{{\tablename\ \thetable{} -- continued from previous page}} \\
\hline Cation Nr. & Cation    & Anion Nr. & Anion                                     \\ \hline 
\endhead

\hline
\endfoot

\hline
\endlastfoot

1  & Tetrabutylammonium  & 1  & Acetate                 \\
2  & Tetrapropylammonium & 2  & Thiocyanate             \\
3  & Tetraethylammonium  & 3  & Carbonate               \\
4  & Tetramethylammonium & 4  & Formate                 \\
5  & Uranyl              & 5  & Nitrite                 \\
6  & Silver(I)           & 6  & Dihydrogen phosphate    \\
7  & Barium (II)         & 7  & Hydrogen phosphate      \\
8  & Calcium(II)         & 8  & Phosphate               \\
9  & Cadmiun(II)         & 9  & Peroxydisulfate         \\
10 & Cobalt(II)          & 10 & Sulfate                 \\
11 & Chromium(III)       & 11 & Arsenate                \\
12 & Cesium(I)           & 12 & Dichromate              \\
13 & Copper(II)          & 13 & Chromate                \\
14 & Dysprosium(III)     & 14 & Nitrate                 \\
15 & Erbium(III)         & 15 & Perrhenate              \\
16 & Europium(III)       & 16 & Selenate                \\
17 & Gadolinium(III)     & 17 & Pertechnetate           \\
18 & Proton              & 18 & Bromide                 \\
19 & Holmium(III)        & 19 & Hexacyanocobaltate(III) \\
20 & Potassium(I)        & 20 & Chloride                \\
21 & Lanthanum(III)      & 21 & Fluoride                \\
22 & Lithium(I)          & 22 & Iodide                  \\
23 & Lutetium(III)       & 23 & Bromate                 \\
24 & Magnesium(II)       & 24 & Chlorate                \\
25 & Manganese(II)       & 25 & Perchlorate             \\
26 & Ammonium            & 26 & Iodate                  \\
27 & Sodium(I)           & 27 & Hydroxide               \\
28 & Neodymium(III)      &    &                         \\
29 & Nickel(II)          &    &                         \\
30 & Praseodymium(III)   &    &                         \\
31 & Rubidium(I)         &    &                         \\
32 & Samarium(III)       &    &                         \\
33 & Strontium(II)       &    &                         \\
34 & Terbium(III)        &    &                         \\
35 & Thorium(IV)         &    &                         \\
36 & Thallium(I)         &    &                         \\
37 & Thulium(III)        &    &                         \\
38 & Yttrium(III)        &    &                         \\
39 & Ytterbium(III)      &    &                         \\
40 & Zinc(II)            &    &

\label{tab:ions_simoes}
\end{longtable}

\end{spacing}
\endgroup










\section{Database Consolidation}

\begin{figure}
    \centering
    \includegraphics[width=\textwidth]{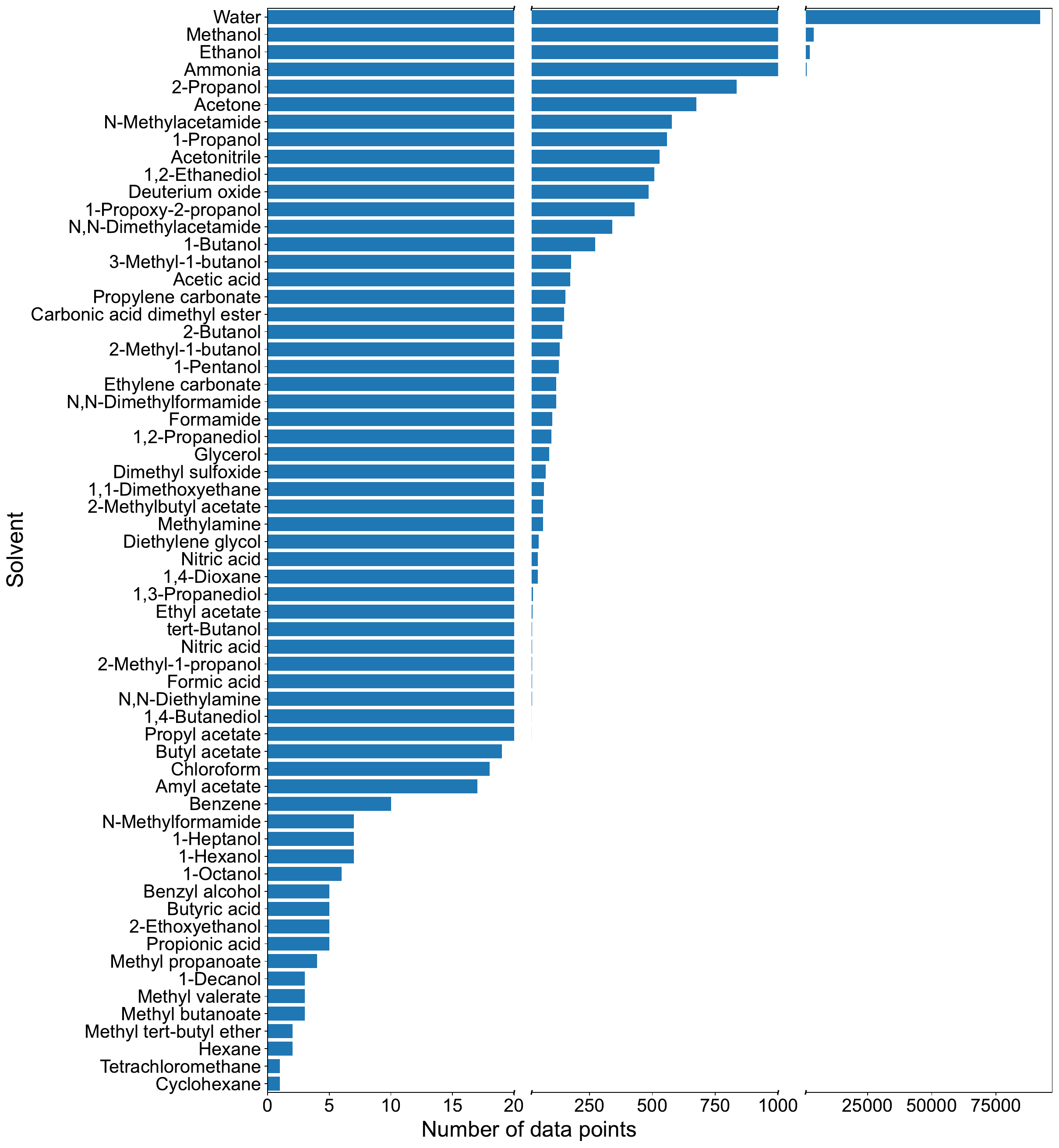}
       \caption{Bar plot representing the number of experimental data points for activity data in binary electrolyte solutions (one electrolyte and one solvent) available in the DDB\cite{ddb} for different solvents.}
    \label{fig:solvents}
\end{figure}

Raw activity data for binary systems containing one electrolyte and one solvent were obtained from the "ELE" (Vapor-Liquid Equilibria of Electrolyte Systems) database of the Dortmund Databank (DDB) 2026\cite{ddb}. The DDB features electrolyte solution activity data for various solvents, as shown in Figure \ref{fig:solvents}.

\begin{figure}
    \centering
    \includegraphics[width=\textwidth]{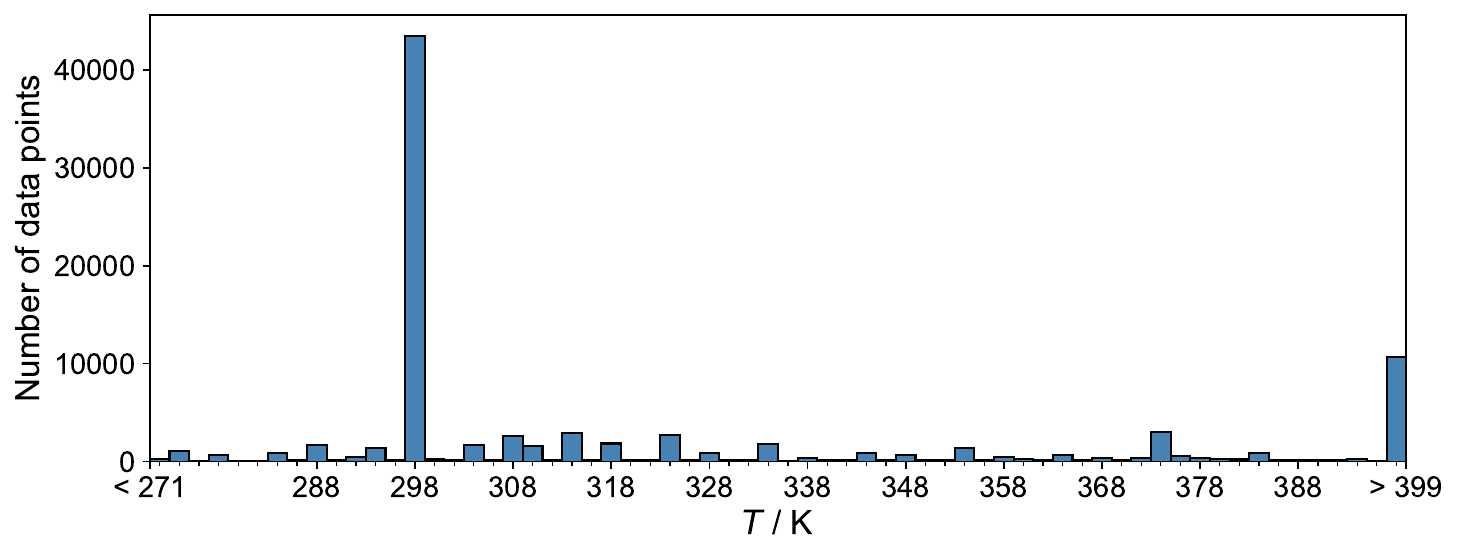}
       \caption{Histogram representing the number of experimental data points for activity data in aqueous solutions of single electrolytes available in the DDB\cite{ddb} for different temperatures.}
    \label{fig:temps}
\end{figure}

As can be seen in Figure \ref{fig:solvents}, the vast majority of the literature data is available for aqueous systems, i.e., systems with water as the solvent. A substantial amount of data is also available for the solvents methanol, ethanol, and ammonia. In contrast, the number of data points for all other solvents is at least two orders of magnitude smaller. Since water is the most common solvent based on the available data points and the Bromley parameters are solvent-specific \cite{Bromley1973}, we restricted the dataset for this study to aqueous systems. With this restriction in place, Figure \ref{fig:temps} shows a histogram of the number of data points as a function of temperature.

As shown in Figure \ref{fig:temps}, most data points are available at $T=298\pm 1$~K. Bromley \cite{Bromley1973} proposed a correlation for the temperature dependence of the $B_{CA}$ parameters using four temperature-independent adjustable parameters for each $B_{CA}$, i.e., for each electrolyte. Since fitting these parameters accurately would, firstly, require a substantial amount of data for each electrolyte, and, secondly, substantially increase the complexity of the developed hybrid model, potentially reducing its ability to generalize to unstudied electrolytes, the present study was restricted to data at $T=298\pm 1$~K, which is also the temperature for which the Bromley model was originally developed.

Furthermore, the dataset was restricted to systems for which either mean ionic activity coefficients $\bar{\gamma}_{\pm}$ and/or osmotic coefficients $\phi$ were available, as well as for which the data were reported using the molality $\tilde{m}$ as the concentration measure. Data reported in terms of molarity were excluded, as conversion to molality requires knowledge of the solution density, which is generally unavailable. Likewise, vapor pressure data were not considered because vapor pressure measurements usually lack sufficient accuracy to calculate exact activity data. In fact, restricting the dataset to molality-based data did not reduce the number of electrolytes it covered.

Since the raw data points were directly adopted from the DDB, it is difficult to provide a reliable estimate of their uncertainty. Many literature sources do not report uncertainties for measured $\phi$ or $\tilde{\gamma}_{\pm}$, and those that do typically state values in the range of 1–5\%. Such estimates appear optimistic, as direct comparisons between different sources, where possible, often revealed deviations of up to 10\%.

\section{Hyperparameter optimization}

\begin{figure}[H]
\centering
    \begin{subfigure}{.49\textwidth}
    \centering
    \includegraphics[width=1\textwidth]{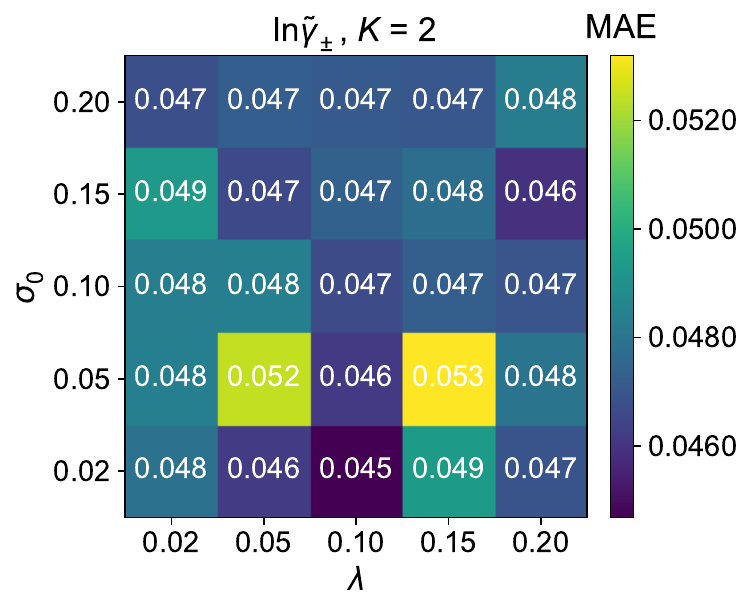}
    \end{subfigure}
    \begin{subfigure}{.49\textwidth}
    \centering
    \includegraphics[width=1\textwidth]{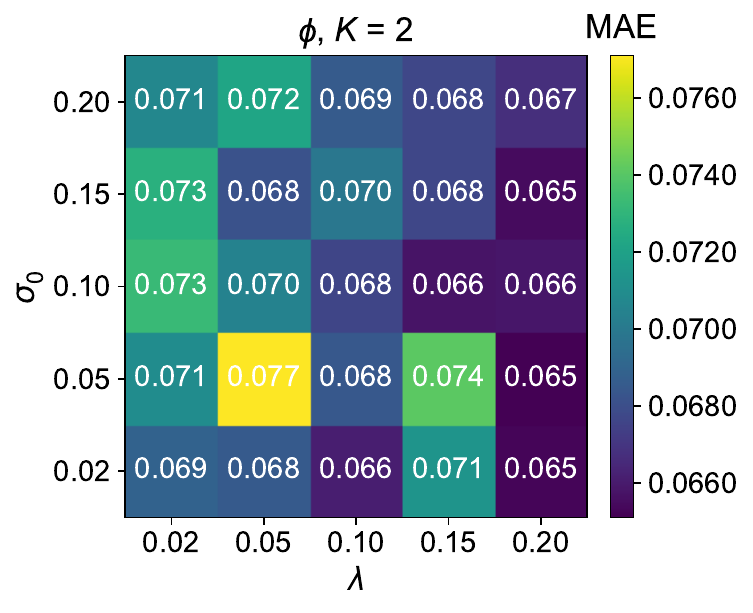}
    \end{subfigure}
    \begin{subfigure}{.49\textwidth}
    \centering
    \includegraphics[width=1\textwidth]{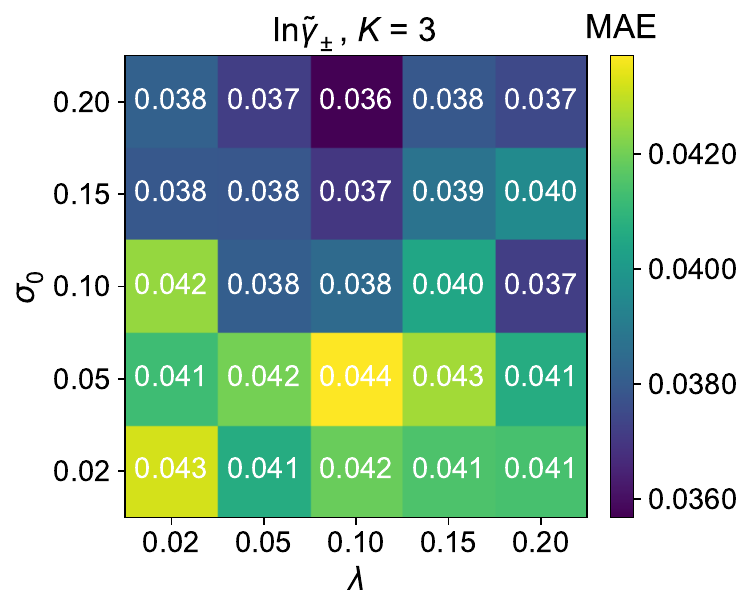}
    \end{subfigure}
    \begin{subfigure}{.49\textwidth}
    \centering
    \includegraphics[width=1\textwidth]{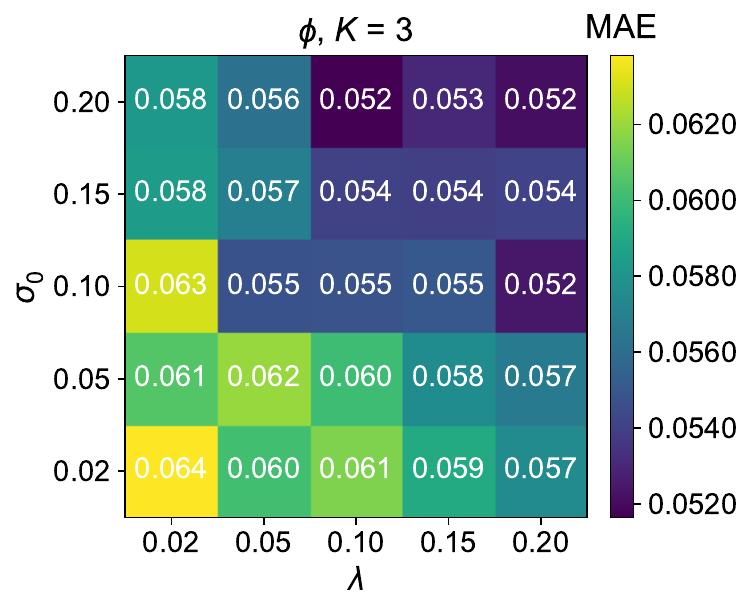}
    \end{subfigure}
    \begin{subfigure}{.49\textwidth}
    \centering
    \includegraphics[width=1\textwidth]{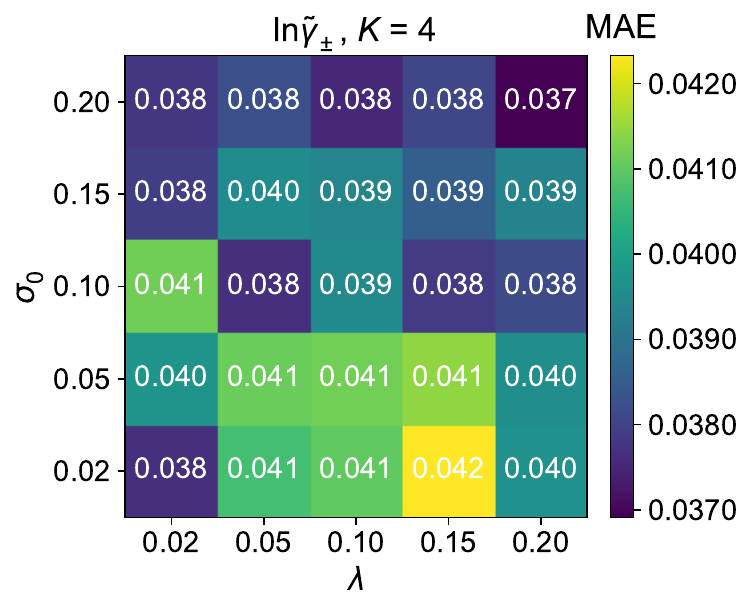}
    \end{subfigure}
    \begin{subfigure}{.49\textwidth}
    \centering
    \includegraphics[width=1\textwidth]{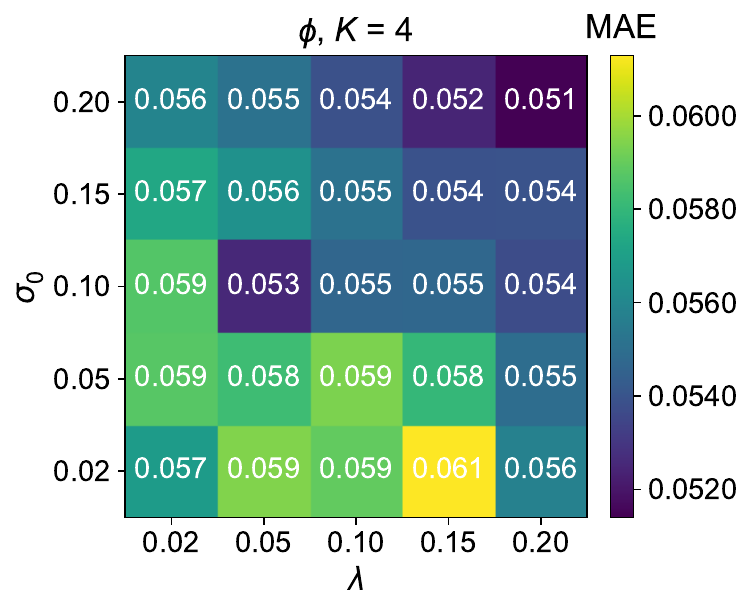}
    \end{subfigure}
    \caption{Prediction MAE for $\ln\tilde\gamma_\pm$ and $\phi$ for different values of the hyperparameters $K$, $\sigma_0$, and $\lambda$ calculated by leave-one-electrolyte-out analysis based on the reduced data set, cf. Figure \ref{fig:database} in the manuscript.}
    \label{fig:hyperparams}
\end{figure}

Figure \ref{fig:hyperparams} shows the prediction error for various combinations of the model hyperparameters $K$, $\sigma_0$, and $\lambda$ evaluated by leave-one-electrolyte-out analysis in the form of mean absolute error (MAE) for $\ln\tilde\gamma_\pm$ and $\phi$.

Figure \ref{fig:hyperparams} shows that setting $K>2$ leads to substantially lower MAE than $K=2$; however, there is only marginal change when increasing $K$ beyond 3. For the $K=2$ model that was used to compare our Bromley-MCM to the ion-specific Bromley model\cite{Bromley1973} (IBM), we chose to set $\sigma_0=0.02$ and $\lambda=0.10$ since those hyperparameters lead to the lowest prediction MAE at $K=2$ for both $\ln\tilde{\gamma}_\pm$ and $\phi$. Meanwhile, for the $K=3$ model, we chose to set $\sigma_0=0.20$ and $\lambda=0.10$ since those hyperparameters lead to the lowest prediction MAE at $K=3$ for $\ln\tilde{\gamma}_\pm$. We consider the marginal improvement for $\phi$ prediction at $K=4$ not sufficient to warrant further increasing the model complexity.

\section{Further Results}

\paragraph{Comparison of original and refitted IBM}

\begin{figure}[H]
    \centering
    \includegraphics[width=0.6\textwidth]{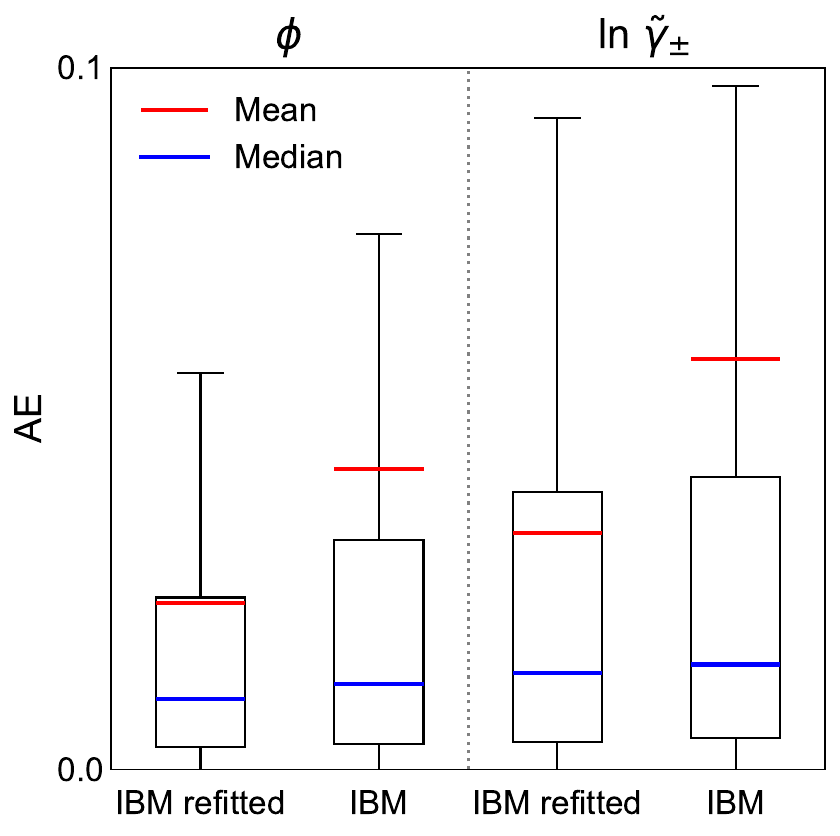}
       \caption{Box plots of the $\ln \tilde{\gamma}_{\pm}$ and $\phi$ fit errors of the original IBM \cite{Bromley1973} and the refitted IBM, fitted on the dataset used in this work. Boxes represent interquartile range (IQR), and whiskers denote 1.5 IQR.}
    \label{fig:ibmoldnew}
\end{figure}

Figure \ref{fig:ibmoldnew} shows the errors for $\ln \tilde{\gamma}_{\pm}$ and $\phi$ of the original IBM\cite{Bromley1973} and the refitted IBM, updated with the data used in this work, on the original IBM horizon in the form of a box plot.

Figure \ref{fig:ibmoldnew} shows that updating the IBM with new experimental data substantially improves the fitting accuracy of the model. We thus consider only the refitted version in the following and the main section of this work, and refer to it simply as IBM. 

\paragraph{Comparison of IBM and Bromley-MCM ($K=3$)}

\begin{figure}[H]
    \centering
    \includegraphics[width=0.6\textwidth]{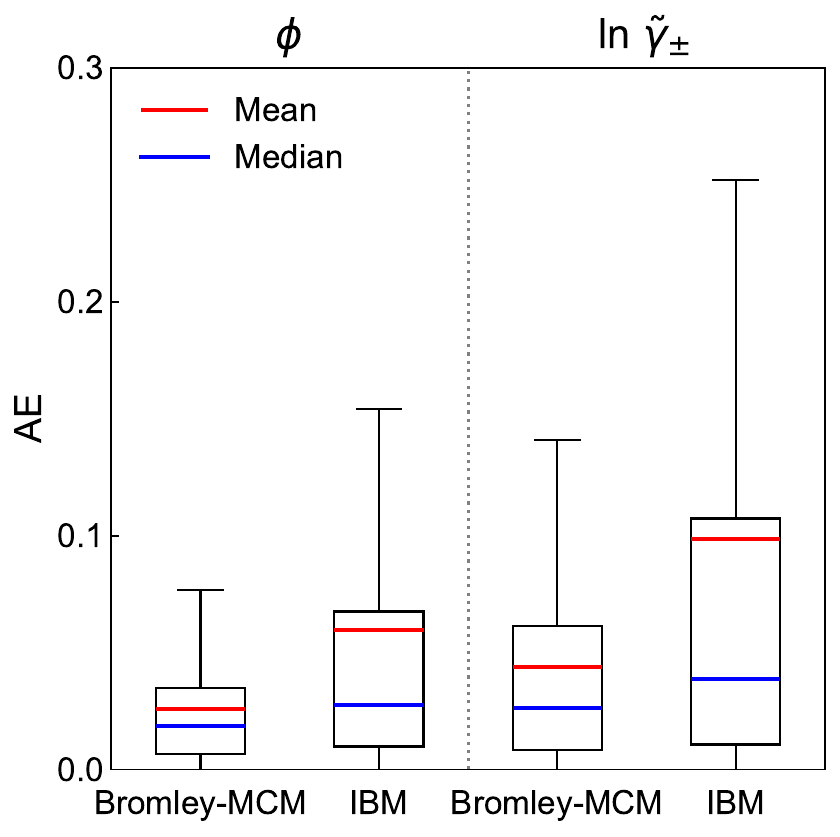}
       \caption{Box plot of the $\ln \tilde{\gamma}_{\pm}$ and $\phi$ fit errors on the reduced database of IBM and Bromley-MCM. Boxes represent interquartile range (IQR), and whiskers denote 1.5 IQR.}
    \label{fig:ibmfits}
\end{figure}

Figure \ref{fig:ibmfits} shows the residuals of the fit of $\ln \tilde{\gamma}_{\pm}$ and $\phi$ for IBM and Bromley-MCM on the reduced data set (Figure \ref{fig:database}) in the form of a box plot. 

\paragraph{Comparison of IBM and Bromley-MCM ($K=2$)}

\begin{figure}[H]
    \centering
    \includegraphics[width=0.6\textwidth]{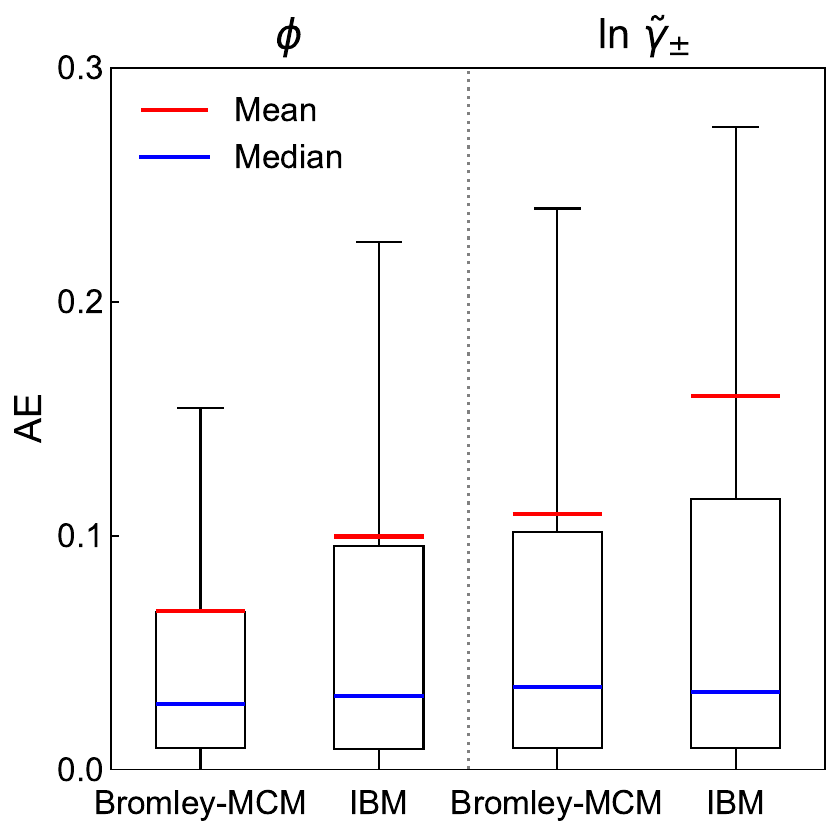}
    \caption{Box plot of the $\ln \tilde{\gamma}_{\pm}$ and $\phi$ prediction errors on the reduced database for IBM and Bromley-MCM (with $K=2$) evaluated using leave-one-electrolyte-out analysis. Boxes represent interquartile range (IQR), and whiskers denote 1.5 IQR.}
    \label{fig:ibmvsbromleyk2}
\end{figure}

Figure \ref{fig:ibmvsbromleyk2} shows a box plot of the prediction errors of $\ln \tilde{\gamma}_{\pm}$ and $\phi$ for IBM and Bromley-MCM using $K=2$ on the reduced data set evaluated by leave-one-electrolyte-out analysis.

The Figures \ref{fig:ibmfits} - \ref{fig:ibmvsbromleyk2} show that our new Bromley-MCM outperforms the existing IBM model\cite{Bromley1973} in all metrics, confirming our conclusions from the main section. 

\begin{figure}[H]
    \captionsetup[subfigure]{
        justification=centering,
        singlelinecheck=false,
        margin={4mm,-4mm}
    }
    \centering
    \begin{subfigure}{.49\textwidth}
    \centering
    \caption*{Seen electrolytes}
    \includegraphics[width=1\textwidth]{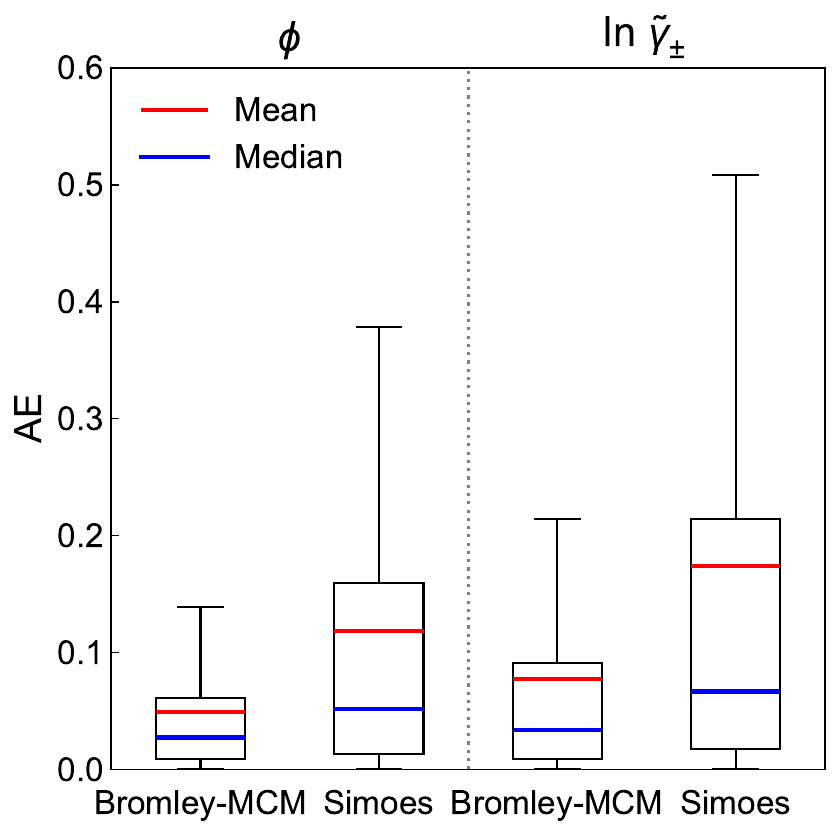}
    \end{subfigure}
    \begin{subfigure}{.49\textwidth}
    \centering
    \caption*{Unseen electrolytes}
    \includegraphics[width=1\textwidth]{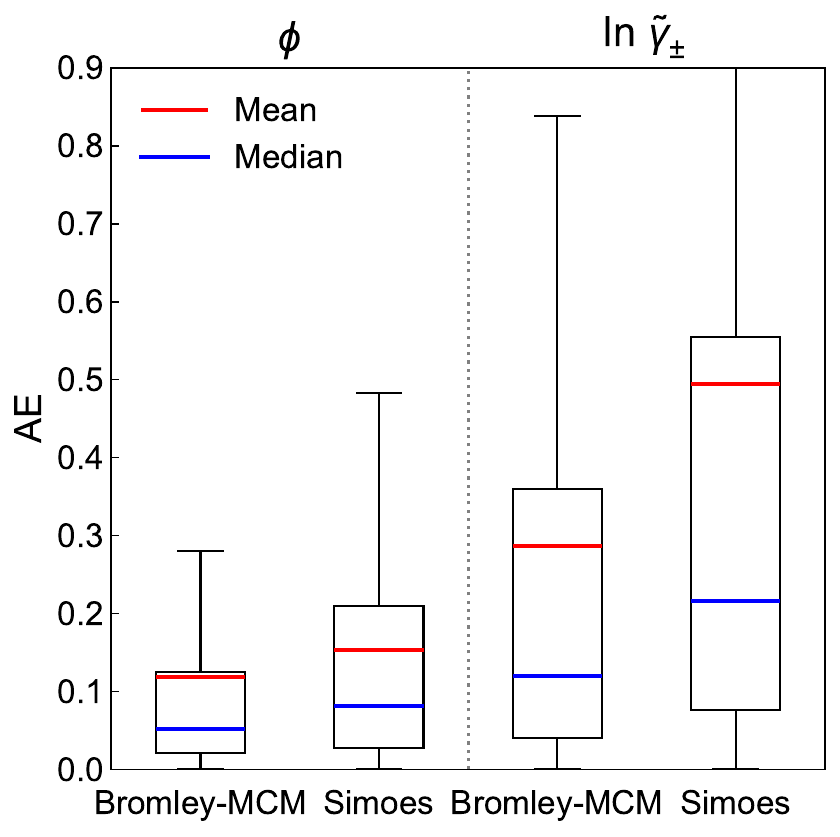}
    \end{subfigure}
    \caption{Box plot of the $\ln \tilde{\gamma}_{\pm}$ and $\phi$ prediction errors on the Simoes horizon for Bromley-MCM and the Simoes model, thereby distinguishing between electrolytes seen by the Simoes model during the training (left) and unseen ones (right). Results of Bromley-MCM are always for unseen electrolytes. Boxes represent IQR, and whiskers denote 1.5 IQR.}
    \label{fig:simoesvsbromley2}
\end{figure}

Figure \ref{fig:simoesvsbromley2} shows a box plot of the prediction errors of $\ln \tilde{\gamma}_{\pm}$ and $\phi$ for Bromley-MCM and the Simoes model, thereby distinguishing between the electrolytes for which data were used during the training of the Simoes model (left) and electrolytes for which no data were used during the training of the Simoes model (right). The errors reported for Bromley-MCM are, in both cases, for truly unseen electrolytes, since this model was evaluated using leave-one-electrolyte-out analysis.

Bromley-MCM significantly outperforms the Simoes model in both cases, indicating that Bromley-MCM's capability to extrapolate to unseen electrolytes is superior to that of the Simoes model. Furthermore, even for electrolytes observed during training of the Simoes model, Bromley-MCM yields, on average, lower errors than the fitted Simoes model.

\section{Data Availability of Osmotic and Activity Coefficients}

\begin{figure}[H]
\centering
    \captionsetup[subfigure]{
        justification=centering,
        singlelinecheck=false,
        margin={4mm,-4mm}
    }
    \begin{subfigure}{.49\textwidth}
    \centering
    \caption*{$\tilde{\gamma}_{\pm}$}
    \includegraphics[width=1\textwidth]{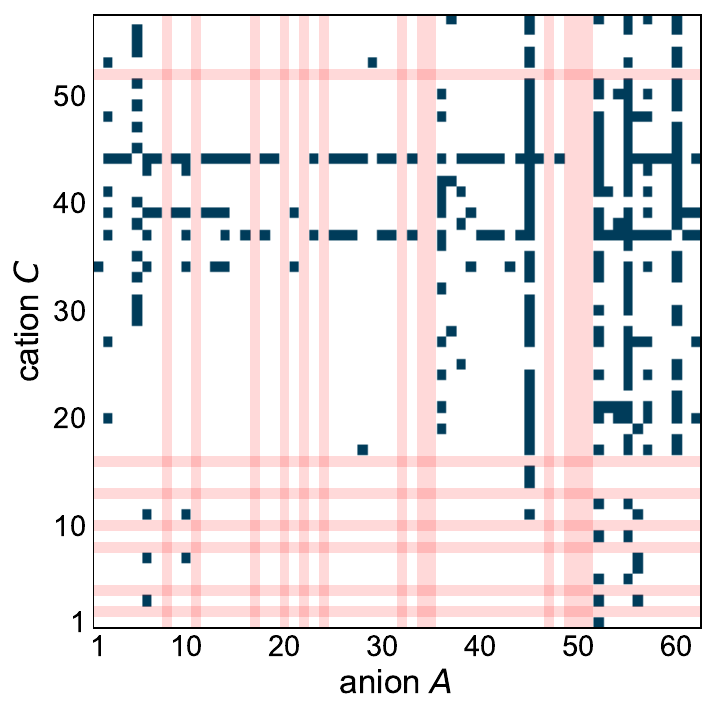}
    \end{subfigure}
    \begin{subfigure}{.49\textwidth}
    \centering
    \caption*{$\phi$}
    \includegraphics[width=1\textwidth]{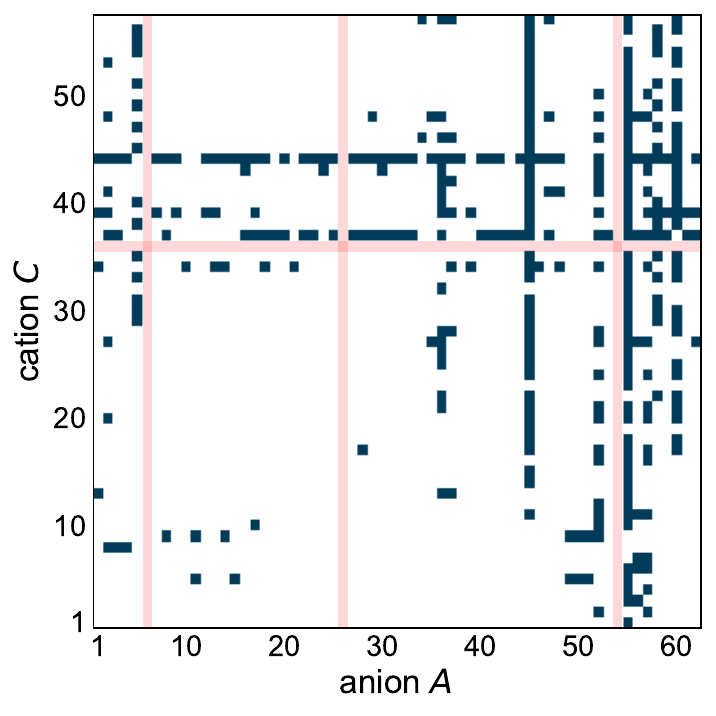}
    \end{subfigure}
    \caption{Experimental data availability for $\tilde{\gamma}_{\pm}$ (left) and $\phi$ (right) for cation-anion pairs at $T=298\pm 1$~K in aqueous systems in the reduced database used in the present work. For the numbers identifying the cations and anions, see Table \ref{tab:ions_reduced}. White cells indicate missing experimental data for specific electrolytes. Red rows and columns denote cations and anions, respectively, with missing $\tilde{\gamma}_{\pm}$ (left) or $\phi$ (right) data.}
    \label{fig:availability_gamma_osm}
\end{figure}

Figure \ref{fig:availability_gamma_osm} shows the availability of experimental data for cation-anion pairs in the reduced database, thereby distinguishing between available data for $\tilde{\gamma}_{\pm}$ and for $\phi$.

Figure \ref{fig:availability_gamma_osm} shows that for many cations and anions no $\tilde{\gamma}_{\pm}$ data are available. Similarly, for three anions and one cation, no $\phi$ data are available. By using both $\tilde{\gamma}_{\pm}$ and $\phi$ data during development and training of the new model, its range of applicability could therefore be increased.
\bibliography{achemso-demo}